\documentclass[10pt,journal,compsoc]{IEEEtran}
\usepackage{graphicx}
\usepackage{booktabs}
\usepackage{bm}
\usepackage{mathtools}
\usepackage{xcolor,colortbl}
\usepackage{hhline}
\usepackage{amsmath,amssymb}
\usepackage{multirow}
\usepackage{enumitem}
\usepackage[pagebackref,breaklinks,colorlinks]{hyperref}

\definecolor{myGreen}{RGB}{0,114,0}
\definecolor{Gray}{gray}{0.95}
\definecolor{turquoise}{RGB}{64,224,208}
\definecolor{pink1}{RGB}{247,146,248}
\definecolor{cyan1}{RGB}{145,253,254}

\ifCLASSOPTIONcompsoc
  \usepackage[nocompress]{cite}
\else
  \usepackage{cite}
\fi
\ifCLASSINFOpdf
\else
\fi

\def\ie{\emph{i.e.}}

\begin{document}
%
\title{Optical Flow Estimation in 360$^\circ$ Videos: \\
Dataset, Model and Application}

\author{Bin Duan, Keshav Bhandari, Gaowen Liu and Yan Yan
\IEEEcompsocitemizethanks{
\IEEEcompsocthanksitem B. Duan and Y. Yan are with Department of Computer Science, Illinois Tech, Chicago, IL, USA 60616.\protect\\
Contact: bduan2@hawk.iit.edu; yyan34@iit.edu
\IEEEcompsocthanksitem K. Bhandari was with Texas State University, present with Tesla Inc, USA.
\IEEEcompsocthanksitem G. Liu is with Cisco Research, USA.
}
}

%
%


\markboth{Journal of \LaTeX\ Class Files, 2023 - Preprint}%
{Duan \MakeLowercase{\textit{et al.}}: Optical Flow Estimation in 360$^\circ$ Videos: Dataset, Model and Application}

%



\IEEEtitleabstractindextext{%
\begin{abstract}
\label{sec:abstract}
Optical flow estimation has been a long-lasting and fundamental problem in the computer vision community. However, despite the advances of optical flow estimation in perspective videos, the 360$^\circ$ videos counterpart remains in its infancy, primarily due to the shortage of benchmark datasets and the failure to accommodate the omnidirectional nature of 360$^\circ$ videos. We propose the first perceptually realistic 360$^\circ$ filed-of-view video benchmark dataset, namely FLOW360, with 40 different videos and 4,000 video frames. We then conduct comprehensive characteristic analysis and extensive comparisons with existing datasets, manifesting FLOW360's perceptual realism, uniqueness, and diversity. Moreover, we present a novel \textbf{S}iamese representation \textbf{L}earning framework for \textbf{O}mnidirectional \textbf{F}low (SLOF) estimation, which is trained in a contrastive manner via a hybrid loss that combines siamese contrastive and optical flow losses. By training the model on random rotations of the input omnidirectional frames, our proposed contrastive scheme accommodates the omnidirectional nature of optical flow estimation in 360$^\circ$ videos, resulting in significantly reduced prediction errors. The learning scheme is further proven to be efficient by expanding our siamese learning scheme and omnidirectional optical flow estimation to the egocentric activity recognition task, where the classification accuracy is boosted up to $\sim$26\%. To summarize, we study the optical flow estimation in 360$^\circ$ videos problem from perspectives of the benchmark dataset, learning model, and also practical application. The FLOW360 dataset and code are available at \url{https://siamlof.github.io}.

\end{abstract}

}

\maketitle

\IEEEdisplaynontitleabstractindextext

%
\IEEEpeerreviewmaketitle

\section{Introduction}
\label{sec:intro}
\IEEEPARstart{O}ptical flow estimation, as a long-lasting and fundamental problem in the computer vision community, has been studied over decades where early works~\cite{lucas, horn} can date back to 80\textit{s}. Nowadays, with the advances of cutting-edge 360$^\circ$\footnote{360$^\circ$ and omnidirectional are using interchangeably in this paper.} sensors and cameras, 360$^\circ$ videos pose a great challenge for optical flow estimation due to most studies being limited to perspective videos, not only benchmark datasets but also methodologies. In fact, the insufficiency of such resources holds back the evolution of optical flow estimation in 360$^\circ$ videos, resulting in surprisingly little work among the computer vision community.

Before the era of modern deep learning, traditional optical flow estimation methods relied on hand-crafted feature designs~\cite{pixel1,pixel2,pixel3}, energy-based optimizations~\cite{energy1,energy2,energy3} and variational approaches~\cite{feature1,variational1,feature2}. Although deep learning-based approaches~\cite{gma,raft,maskflownet,pwc,flownet2,liteflownet} have shown impressive advantages over classical approaches, most works are specially tailored for perspective videos. The easy access to such perspective datasets~\cite{sintel,kitti,kitti2012,kitti2015,middlebury} further stimulates advances of these modern deep learning-based approaches for perspective optical flow estimation, which is not the case for omnidirectional flow estimation research. The disproportional numbers of omnidirectional research is an implicit indicator that omnidirectional optical flow estimation research remains in its infancy.

To stimulate omnidirectional optical flow estimation research, a high-quality benchmark dataset is in high demand. This need for a benchmark dataset brings up the first challenge: there is no such reliable, \ie, perceptually natural and complex 360$^\circ$ video dataset in the literature for omnidirectional optical flow estimation. This paper addresses the first challenge of reliable benchmark dataset shortage by proposing a new dataset named FLOW360. To the best of our knowledge, this is the very first perceptually natural-synthetic 360$^\circ$ video dataset for omnidirectional flow estimation. Unlike existing datasets, our proposed dataset advances mainly in two aspects, \ie, full 360$^\circ$ field of view (FOV) and high perceptual realism. Comparatively, OmniFlow\cite{omniflow} dataset only has 180$^\circ$ FOV failing to address the omnidirectional nature, while the dataset proposed in OmniFlowNet\cite{omniflownet} lacks perceptual realism in scene and motion. We conduct comprehensive characteristic analysis on the proposed dataset and extensive comparisons with existing datasets, which further manifest FLOW360's perceptual realism, uniqueness, and diversity. Given the success of perspective optical flow datasets such as~\cite{middlebury,sintel,kitti} facilitating researchers' investigation on perspective optical flow estimation methods~\cite{liteflownet,flownet,flownet2,maskflownet,raft}, the availability of such omnidirectional video dataset is essential to advance this particular field. Moreover, it is worth noting that FLOW360 can also be used in various other areas, such as continuous flow estimation in 3-frame settings with forward/backward consistency~\cite{selflfow,unflow,mirrorflow}, depth estimation~\cite{omnidepth1,omnidepth2} and normal map estimation~\cite{normal}.

Another challenge of omnidirectional optical flow estimation is that current perspective video-based deep networks fail to accommodate the nature of 360$^\circ$ videos. These perspective methods inevitably require fine-tuning due to the presence of radial distortion~\cite{radial} in 360$^\circ$ videos, which is caused by projecting 360$^\circ$ videos (spherical) to an equirectangular plane. This fine-tuning is onerous and requires tricky transformative techniques to adapt the radial distortion~\cite{ktn,tangent}, such as spherical convolution~\cite{ktn,spherical1,spherical2,spherical3}, spectral convolution~\cite{spectral1,spectral2} and tangent convolution \cite{tangent}. Generally speaking, these methods require significant modifications of convolution layers, \ie, from perspective kernel to omnidirectional-aware kernel. Not only they require immense effort to redo layer-wise architecture design, but the dramatic changes of underlying kernels also restrict the transition from fruitful perspective studies to the less-explored omnidirectional field. Other than modifying the underlying convolutional kernel, an intuitive solution is to fine-tune perspective-based deep networks under omnidirectional supervision. However, this brute-force migration of perspective-based networks often requires enormous data and still leads to significant performance degradation~\cite{revisiting}. 

Instead of switching to new convolution layers and also to ease the amount of supervision, we design a novel \textbf{S}iamese representation \textbf{L}earning for \textbf{O}mnidirectional \textbf{F}low (SLOF) framework, which leverages the rotation-invariant property of omnidirectional videos to address the radial distortion problem. The term rotation-invariant here implies that we can recover the original projection of 360$^\circ$ videos given the reverse rotation of the applied projection. The good thing about this rotation-invariant property is that we can project omnidirectional videos on any plane by rotating the spherical videos on three different axes $XYZ$, namely \textit{pitch}, \textit{roll} and \textit{yaw}, while these rotation operations still preserve the overall information. That is, any projected planar representation can be converted back to the spherical representation since we define the rotation beforehand. This rotation-invariant property motivates our design of the proposed SLOF, where we estimate the omnidirectional optical flow from a pair of consecutive frames and their rotated counterparts, assuming that the representations of these two cases are similar enough so that it can generate nearly identical optical flow in the spherical domain. We also design and compare different combinations of rotational strategies and derive guidelines for selecting the most effective augmentation scheme.

When it comes to practical applications such as egocentric activity recognition, 360$^\circ$ videos impose a challenge of significant performance drop due to random FOV projections. For example, since the trained model typically only saw one projected planar videos, it fails to capture an overall omnidirectional representation for prediction. This paper introduces our siamese representation learning framework in the egocentric activity recognition task. We propose Vision Transformer 360 (VIT360) to address the aforementioned issue. Compared to accuracy-centric convolution-based architecture trained on fixed FOV, VIT360 is rotational-invariant and therefore does not suffer from performance degradation when the field-of-view projection is randomized. Besides, VIT360 provides the foundation to leverage motion inference techniques from off-the-shelf optical flow architecture, where researchers are free to choose their favorite networks. We hope this egocentric activity recognition task showcases that omnidirectional learning can significantly benefit from incorporating the rotation-invariant property in the network architecture design for practical application.

We summarize our main contribution as follows: (1) we introduce FLOW360, a new optical flow dataset for omnidirectional videos, to fill the need for a benchmark dataset to advance the omnidirectional flow estimation field. Comprehensive characteristic analysis of FLOW360 and extensive comparisons with existing datasets manifests FLOW360's perceptual realism, uniqueness, and diversity; (2) We propose SLOF, a novel framework for optical flow estimation in omnidirectional videos, to mitigate the cumbersome framework adjustments for omnidirectional flow estimation, and well-accommodated to the nature of 360$^\circ$ videos; (3) We propose VIT360 and introduce siamese learning into the egocentric activity recognition task to mitigate the performance drop problem. This design of VIT360 tackles the challenges of processing 360$^\circ$ videos with a fully convolution-based framework while learning the rotational invariant representation to keep up the high performance. Overall, the FLOW360 dataset, the SLOF framework, and the VIT364 network, as well as our experimental results, provide a solid foundation for future exploration in this interesting and important omnidirectional research field.

\section{Related Work}
\label{sec:relatedwork}

\noindent \textbf{Optical Flow Datasets.} Perspective datasets such as \cite{barron,middlebury,mccane,real1,real2,spectra1} comprise synthetic image sequences along with the synthetic and hand-crafted optical flow. However, these datasets fall short in terms of perceptual realism and complexities. Even though several optical flow datasets have been published recently in \cite{outdoor,kitti,kitti2012,kitti2015}, they are primarily used in automotive driving scenarios. The other relevant dataset in the literature was Sintel~\cite{sintel}, which provided a bridge to contemporary optical flow estimation and synthetic datasets that can be used in real-world situations.

All datasets, as mentioned earlier, are introduced for perspective videos and thus cannot be used for omnidirectional flow estimation. To address this problem, LiteFlowNet360~\cite{revisiting} on omnidirectional flow estimation was released to augment the Sintel dataset by introducing distortion artifacts for the domain adaptation task. Nevertheless, these augmented datasets are discontinuous around the edges and violate the 360$^\circ$ nature of omnidirectional videos. The closest datasets to ours are OmniFlow~\cite{omniflow} and OmniFlowNet~\cite{omniflownet}. OmniFlow introduced a synthetic 180$^\circ$ field-of-view (FOV) dataset, which is limited to indoor scenes and lacks full 360$^\circ$ FOV.
Similarly, OmniFlowNet introduced a full 360$^\circ$ FOV dataset. However, both datasets lack complexities and evidence for perceptual realism. We show detailed comparisons to OmniFlow, and OmniFlowNet in the dataset characteristic study. Compared to existing datasets in the literature, FLOW360 is the first perceptually natural benchmark 360$^\circ$ dataset and fills the void in current research.

\noindent \textbf{Optical Flow Estimation.} Advancements in optical flow estimation techniques largely rely on the success of data-driven deep learning frameworks. Flownet~\cite{flownet} marked one of the initial adoption of CNN- based deep learning frameworks for optical flow estimation. Several other works~\cite{flownet2,liteflownet,flow0,flow1,flow2,flow3,flow4,flow5} followed the footsteps with improved results. Generally, these networks adopt an encoder-decoder framework to learn optical flow in a coarse-to-fine manner. The current framework RAFT~\cite{raft} has shown improvements with correlation learning.

The methods mentioned above are insufficient on omnidirectional flow field estimation as they are designed and trained for perspective datasets. One initial work~\cite{backproject} on omnidirectional flow estimation was presented as flow estimation by back-projecting image points to the virtually curved retina, thus called back-projection flow. It showed an improvement over classical algorithms. Similarly, another classical approach~\cite{wavelet} relied on spherical wavelet to compute optical flow on omnidirectional videos. However, these methods are limited to classical approaches as they are not relevant in existing deep learning-based approaches. One of the recent works, LiteFlowNet360~\cite{revisiting} tried to compute optical flow on omnidirectional videos using domain adaptation. This method utilized the kernel transformer technique (KTN~\cite{ktn}) to adapt convolution layers on LiteFlowNet~\cite{liteflownet} and learn correct convolution mapping on spherical data. Similarly, OmniFlowNet~\cite{omniflownet} proposed a deep learning-based optical flow estimation technique for omnidirectional videos. The major drawback of these methods is the requirement to adapt convolution layers, which takes a substantial amount of time and makes portability a significant issue. For example, in LiteFlowNet360, each convolution layer in LiteFlowNet was transformed using KTN with additional training and adjustments. Similar to OmniFlowNet, every convolution layer in LiteFlowNet2~\cite{liteflownet2} was transformed using kernel mapping~\cite{kmapping} based on different locations of the spherical image. These techniques incur computational overheads and limit the use of existing architectures. Such approaches demand explicit adaptation of convolution layers, which is hard to maintain when more up-to-date methods are published constantly. Contrary to these methods, we propose a Siamese Representation Learning for Omnidirectional Flow (SLOF) method to learn omnidirectional flow by exploiting existing architectures with designed representation learning objectives, significantly reducing the unnecessary effort of transforming or redesigning the convolution layer.

\noindent \textbf{Action/Activity Recognition.} The convolution-based action/activity recognition framework has been the standard practice in the activity recognition field. Though 2D CNN shows massive success in still image-based applications, incorporating the temporal aspects for the downstream task is challenging because the CNN alone does not account for time. The initial work \cite{karpathy} shows a way to use CNN to incorporate temporal aspects for the video classification task. Similarly, the two-stream architecture \cite{twostream} learns spatio-temporal features from input RGB and optical flow information for video classification. Many other approaches~\cite{lstm1,lstm2, lstm3, lstm4} leverage the recurrent neural network to model long-term dependencies across input video frames. Following the first introduction of 3D CNN based video approach~\cite{first3d}, several other variants of 3D CNN based approaches~\cite{13d,23d,33d,i3d} were proposed. Our transformer-based approach, a hybrid method, takes inspiration from these methods to extract video features. 

Though initially introduced in NLP~\cite{attentionfirst}, the attention mechanism in deep learning has changed several aspects of computer vision research. Ranging from still image-based vision tasks~\cite{transimage} to video action classification task~\cite{videotransformer}, the attention-based mechanism is drawing increasing interest in application areas like the activity recognition domain. The introduction of the paper, ``Attention is all you need''~\cite{attentionneed} marks a significant turning point in transformer-based approaches. This attention framework led to the notable outcome of the transformer-based vision applications as Vision Transformer~\cite{vit} and activity recognition frameworks~\cite{actortransformer, stacktransformer, trear, spacetime}. However, research in egocentric activity recognition in 360$^\circ$ videos is still in its infancy. Inspired by recent advancements in transformer-based vision tasks, we combine the potential of vision transformers with traditional convolution-based techniques for egocentric activity recognition.

Although there are affluent datasets in the egocentric activity recognition literature, most of the published datasets~\cite{kitchen, egohands, beholder, charades, dailyliving} are limited to the perspective field-of-view. The 360$^\circ$ videos based egocentric activity recognition datasets are relatively scarce. EGOK360~\cite{egok360} is a recently published egocentric dataset with 360$^\circ$ field-of-view, which includes two confusing terminologies called activity and actions. The activity is defined as a collection of minor actions, where the actions are more fine-grained motions related to egocentric activities. We perform our experiments on these datasets by focusing only on activity recognition (twelve different classes).

\noindent \textbf{Siamese Representation Learning.} Siamese representation learning is a powerful approach in the unsupervised learning category. Siamese networks have shown great success in different vision-related tasks such as verification~\cite{signature,verification,oneshot} and tracking~\cite {tracking}. A recent approach~\cite{simsiam} in siamese representation learning showed impressive results in unsupervised visual representation learning via exploiting different augmentation views of the same data. They presented their work in pre-training and fine-tuning stages, where the former being the unsupervised representation learning. We use the representation learning scheme on omnidirectional data via rotational augmentations, maximizing the similarity for latent representations and minimizing the flow loss.
\begin{figure*}[!t]
\centering
    \includegraphics[width=0.95\linewidth]{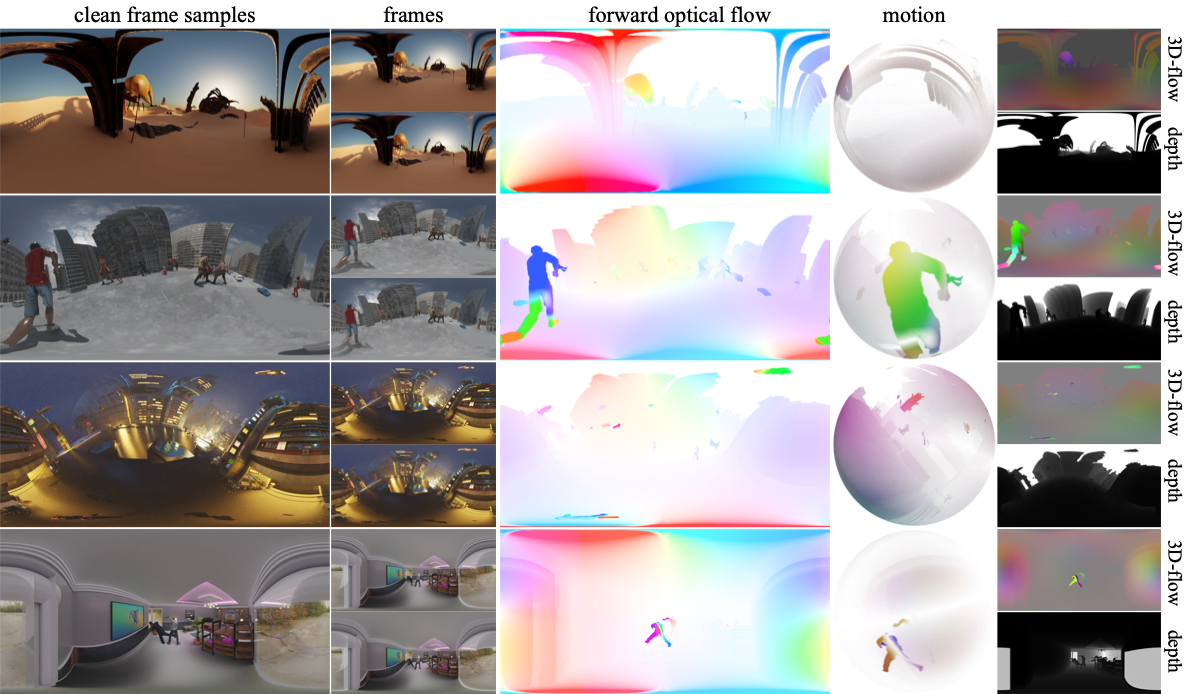}
    \caption{\textbf{The FLOW360 Dataset}. Sample frames (first and second column, respectively) from some of the videos with corresponding forward optical flow and dynamic depth information. Motion in the 3D Sphere (fourth column) is computed by transforming the motion vectors from Equirectangular plane $(\theta, \phi)$ to unit sphere $f(x,y,z)$. Motion in the sphere is represented in RGBA color notation. RGB color representation (as suggested in Middlebury~\cite{middlebury}) is encoded using $(x,y)$ components, and the alpha color is encoded from $z$ of a unit sphere. RGB encoding (fifth column) is an RGB color map of flow in 3D space. \textbf{Note}: flow fields are clipped for better visualization.}
    \label{fig:samples}
\end{figure*}
\section{FLOW360 Dataset}
\label{sec:dataset} 
\begin{figure*}[!t]
\centering
    \includegraphics[width=1\linewidth]{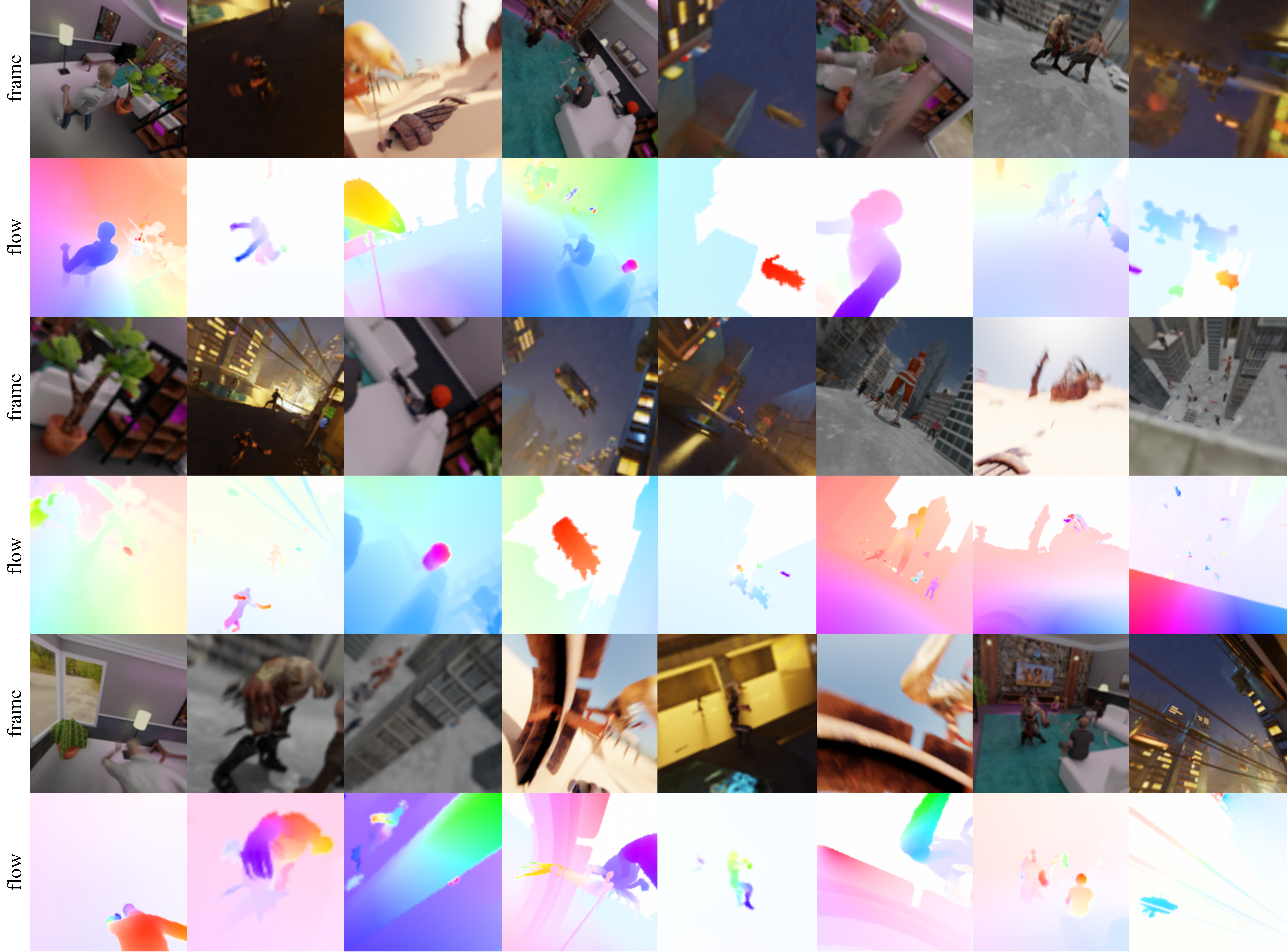}
    \caption{\textbf{Motion and Scene Diversity}. Samples from our FLOW360 Dataset with random projections, showing scene and motion diversity. The FLOW360 dataset has a vast scene consisting of several lighting scenarios, textures, diverse 3D assets, and motion complexity in different regions.}
    \label{fig:pers_flow}
\end{figure*}
\begin{figure}[!t]
\centering
    \includegraphics[width=\linewidth]{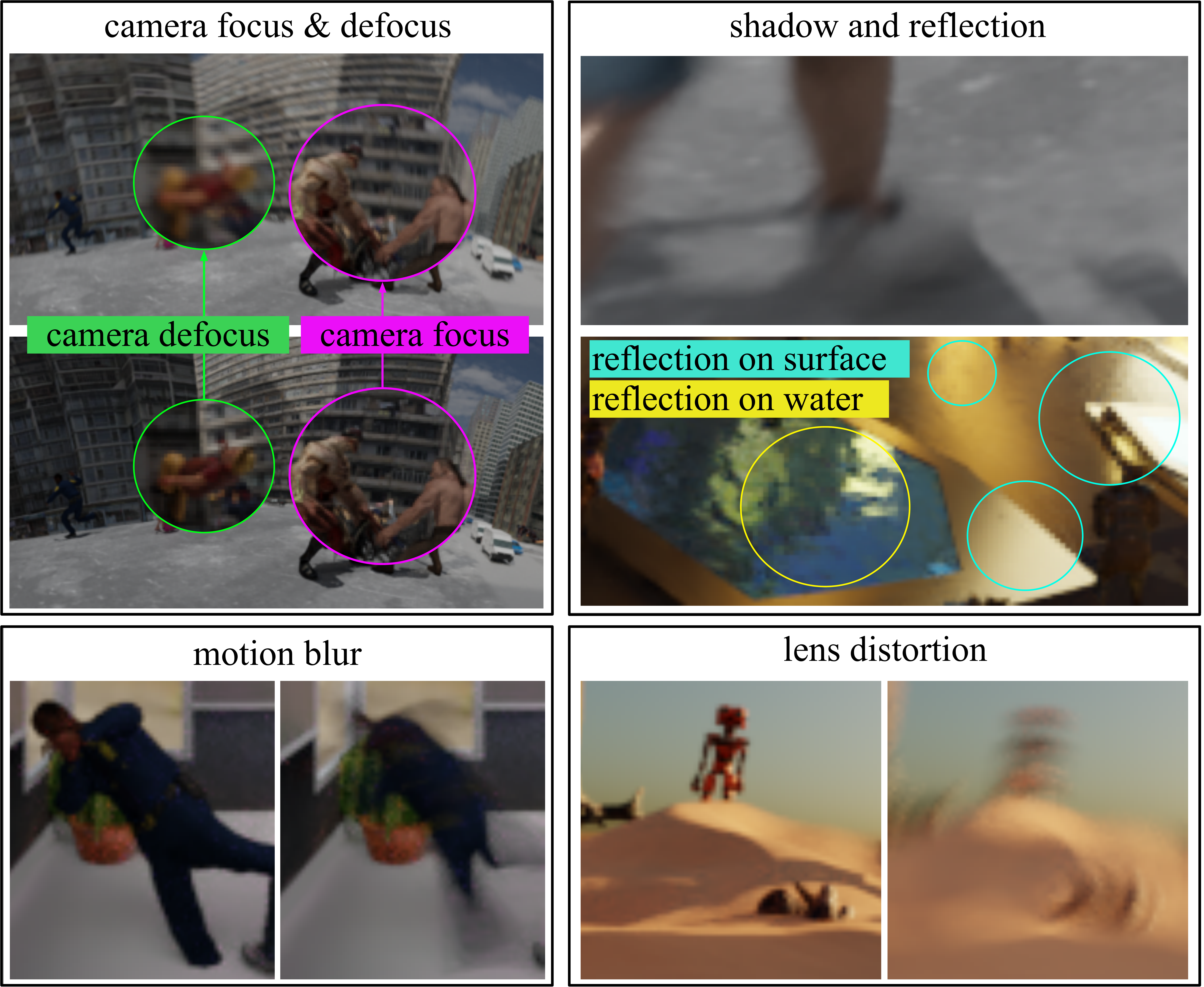}
    \caption{\textbf{Complexity of FLOW360 Dataset}. Frames in FLOW360 Dataset include complex characteristics like camera focus/defocus, motion blur, lens distortion, shadow, and reflections. Our dataset provides ambiance occlusion and environmental effects for a realistic visual appearance.}
    \vspace{-8pt}
    \label{fig:complexity}
\end{figure}
\begin{figure*}[!t]
\centering
    \includegraphics[width=\textwidth]{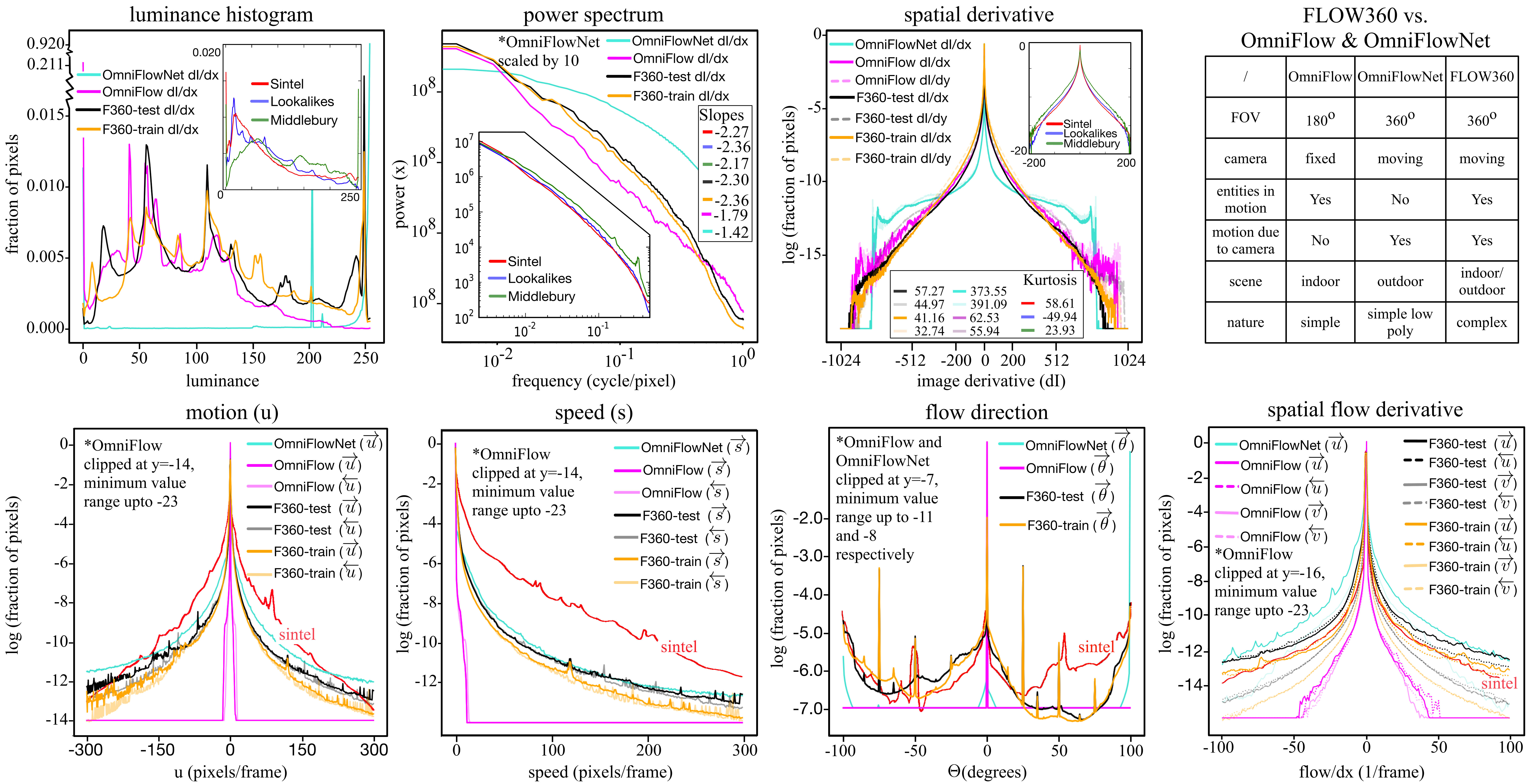}
    \caption{\textbf{Comparision of frames and flow statistics}.  The top row represents the statistics and comparison of the frame with Sintel, Lookalikes, Middlebury, OmniFlow~\cite{omniflow} and OmniFlowNet~\cite{omniflownet}. Bottom row represents flow statistics and comparison with Sintel (\textcolor{red}{red}), OmniFlow (\textcolor{magenta}{magenta}) and OmniFlowNet (\textcolor{turquoise}{turquoise}). The table on the top-right shows a brief comparison of OmniFlow \& OmniFlowNet with the FLOW360 dataset. \textbf{Note}: ($\rightarrow$, $\leftarrow$) represents forward and backward flow fields, respectively.}
    \label{fig:datastat}
\end{figure*}
FLOW360 is an optical flow dataset tailored for 360$^\circ$ videos using Blender~\cite{blender}. This dataset contains naturalistic 360$^\circ$ videos, forward and backward optical flow, and dynamic depth information. The dataset comprises 40 different videos extracted from huge 3D-World `The Room’, `Modern’, `Alien Planet’, and `City Rush’. Due to their size, this 3D-World cannot be rendered at once in a single video. We render several parts of this 3D-World, which provides enough qualitative variation in motion and visual perception like 3D assets, textures, and illuminations. The nature of this large and diverse animated world provides relatively enough diversity to qualify for a standard benchmark dataset. Fig.~\ref{fig:pers_flow} shows some of the examples of motion and scene diversity of FLOW360. Similarly, samples from the dataset of different 3D-World are shown in Fig.~\ref{fig:samples}. We build this 3D-World using publicly available 3D models~\cite{alienplanet,cityrush,modern} and 3D animated characters~\cite{turbosquid,sketchfab,mixamo}. Meanwhile, we adopt Blender~\cite{blender} for additional rigging and animation for the dataset.

FLOW360 contains 40 video clips extracted from different parts of huge 3D-World, `The Room’, `Alien Planet’, `City Rush’, and Modern’. The datasets also contain other information like depth maps and normal fields extracted from the 3D-World. The FLOW360 dataset has 4,000 video frames, 4,000 depth maps, and 3,960 flow fields. We divide the video frames into 2700/1300 train/test split. We render the video frames with the dimension of $(512,1024)$ to save the rendering time. However, FLOW360 can be rendered with higher resolution, as 3D models and Blender add-ons will also be public.

\noindent \textbf{Diversity.} We design FLOW360 datasets to include a diverse situation that resembles the real-world scenario as much as possible. The statistical validity of the datasets in terms of perceptual realism of scene and motion is presented in Fig.~\ref{fig:datastat}. The datasets contain a wide range of motion complexity from smaller to larger displacement, occlusion, motion blur, and similar complexities on the scene using camera focus-defocus, shadow, reflections, and several distortion combinations. As these complexities are quite common in natural videos, the FLOW360 provides similar complexities. Similarly, the datasets cover diverse scenarios like environmental effects, textures, 3D assets, and diverse illuminations. The qualitative presentation of these diversities and complexities are presented in Fig.~\ref{fig:pers_flow} and Fig.~\ref{fig:complexity} respectively. 

\noindent \textbf{Fairness.} The FLOW360 dataset contains custom-tailored animated 360 videos. We plan to release the dataset with the 3D models and our custom Blender add-ons to provide researchers a platform to create their custom optical flow datasets for all kinds of environments (perspective, 180$^\circ$ and 360$^\circ$ FOV). However, the release of 3D world scenes can raise questions regarding fairness. To mitigate this issue, we will perturb certain parts of 3D world scenes and not release any camera information related to the test set.

\noindent \textbf{Flow-generator with Blender Add-ons.} Flow-generator is a custom Blender add-on written for Blender-v2.92. The flow-generator serves two basic purposes. First, it creates a Blender compositor pipeline to collect frames, depth maps and optical flow information. This add-on can also collect additional information, such as normal maps. Second, it sets up a camera configuration for 360$^\circ$ FOV. We will describe the details of the add-ons in the supplementary material.

\noindent\textbf{Render Passes.} We exploit several modern features from Blender-v2.92 like advanced ray-tracing as a render engine along with render passes like vector, normal, depth, mist, and so on to produce realistic 3D scenes. Additionally, we incorporate features like ambient occlusion, motion blur, camera focus/defocus, smooth shading, specular reflection, shadow, and camera distortion to introduce naturalistic complexity (shown in Fig.~\ref{fig:complexity}) in our dataset. Besides optical flow information, the FLOW360 3D-world may be used to collect several other helpful information like depth, normal maps, and semantic segmentation.

\noindent \textbf{Dataset Statistics.} We conduct a comprehensive analysis and compare our dataset with Sintel~\cite{sintel}, Lookalikes (presented in the original Sintel paper to compare the image statistics with the simulated dataset), Middlebury~\cite{middlebury}, OmniFlow~\cite{omniflow} and OmniFlowNet~\cite{omniflownet}. The analysis shown in Fig.~\ref{fig:datastat} shows the image and motion statistics in the top and bottom rows, respectively.

Based on analysis from Sintel, we present frame statistics with three different analysis: luminance histogram, power spectrum, and spatial derivative. For luminance statistics, we convert the frames to gray-scale, $I(x,y)\in[0,255]$ then we compute histograms of gray-scale images across all pixels in the entire dataset. The luminance statistics show the FLOW360 has a similar distribution with the peak in the range between $[0{-}100]$ and decreasing luminosity beyond that range. Similarly, we estimate power spectra from the 2D FFT of the $512{\times}512$ in the center of each frame. We compute the average of these power spectra across all the datasets. We present power spectra analysis separately for the training and test set in this analysis. The power spectra analysis closely resembles the Sintel, Lookalikes, and Middlebury datasets. Based on~\cite{spectra1,spectra2}, the real-world movies exhibit a characteristic of a power spectrum slope around -2, which is equivalent to a $1/f^{2}$ falloff. FLOW360 with the slope $(-2.30, -2.36)$ on test and training split shows such characteristics. We do not claim that FLOW360 is realistic, but it certainly exhibits perceptual similarity with natural movies. The spatial and temporal derivative analysis additionally supports this characteristic. The Kurtosis of frames' spatial derivatives range from 32.74 to 57.27, peaking at zero. This characteristic shows that FLOW360 has a resemblance to natural scenes~\cite{spectra1}.

Regarding the flow field analysis, we directly compare the distribution of motion $u(x,y)$, speed defined as $s(x,y)=\sqrt{u(x,y)^2 + v(x,y)^2}$, flow direction $\Theta(x,y)=\tan^{-1}{(v(x,y)/u(x,y))}$ and spatial flow derivative of $u$ and $v$. The close resemblance of the flow field statistics between Sintel and FLOW360 suggests motion field resemblance with natural movies. Based on these comparisons, FLOW360 exhibits sufficient properties evident enough for its perceptual realism and complexities.

\noindent \textbf{Comparison with OmniFlow and OmniFlowNet.} OmniFlow~\cite{omniflow} presents an omnidirectional flow dataset that is roughly similar to FLOW360. However, the major distinction between these datasets is the FOV. FLOW360 provides immersive 360$^\circ$ FOV, whereas OmniFlow provides only 180$^\circ$ FOV showing FLOW360 compared to OmniFlow is the true omnidirectional dataset. Similarly, OmniFlowNet~\cite{omniflownet} presents synthetic omnidirectional flow dataset with 360$^\circ$ FOV. However, this dataset contains low poly unnatural scenes, which can be explained by relatively larger kurtosis $(373.55,391.09)$, characteristic of a power spectrum and luminance distribution (peaked at 255). The overall statistical analysis reveals FLOW360’s better perceptual realism and diversity.

\noindent \textbf{Applications.} As we mentioned, the FLOW360 dataset contains frames and forward flow fields and includes backward flow fields, depth maps, and 3D-FLOW360 worlds, providing potential for applications like continuous flow-field estimation in 3 frames setting. Besides optical flow estimation, the FLOW360 dataset can be used in other applications such as depth and normal field estimation. Moreover, given 3D-FLOW360 animation data, the researcher can create as many optical flow datasets as needed.
\section{360$^\circ$ Optical Flow Estimation}
\label{sec:method}
\begin{figure*}[!t]
    \centering
    \includegraphics[width=0.75\linewidth]{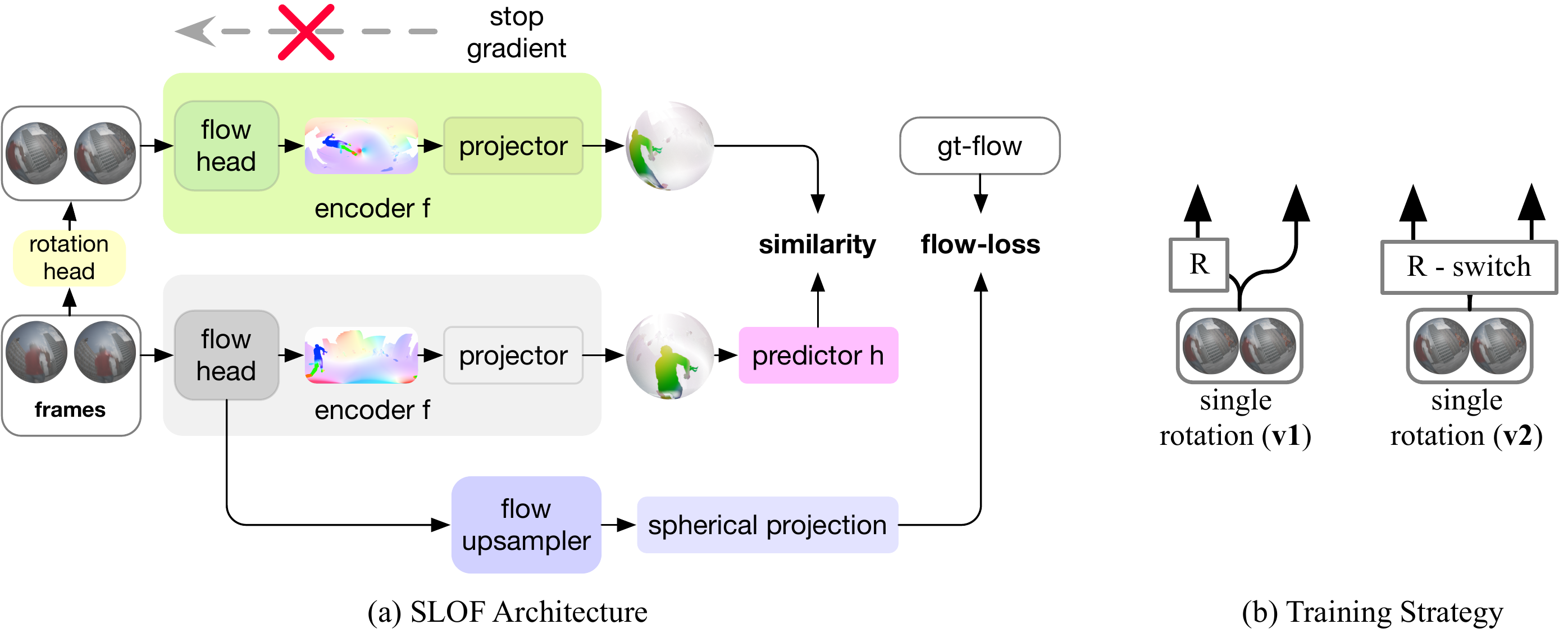}
    \caption{\textbf{Siamese Representation Learning for Omnidirectional Flow (SLOF)}. Pairs of frame sequence (w/ and w/o random rotation) are passed as inputs to encoder \textit{f} (RAFT as a flow head backbone and a standard convolutional projector layer). A predictor layer \textit{h} is an MLP layer. The entire framework is trained by fusing the pretraining and fine-tuning stages to combine the similarity and flow loss in a single stage. The model maximizes the similarity between latent representations of flow information from two streams and minimizes the flow loss. \textbf{Training Strategy (right):} Here, two different arrows~(left, right) represent siamese streams or input pathways to our model. \textbf{v1} and \textbf{v2} (either stream is subjected to rotational augmentation) are similar strategies achieving overall better performance.}
    \label{fig:architecture}
\end{figure*}
\subsection{SLOF for 360$^\circ$ Optical Flow Estimation}
SLOF, as shown in Fig.~\ref{fig:architecture}, is inspired by the recent work on Siamese representation learning~\cite{simsiam}. Since the method we rely on acts as a hub between several methods like contrastive learning, clustering, and siamese networks, it exhibits two special properties required for our case. First, this method has non-collapsing behavior. Here, the term collapsing refers to a situation where an optimizer finds possible minimum -1 similarity loss resulting degenerate solution, characterized by zero standard deviation ($std$) of $l_2$-normalized output $z/||z||_2$ for each channel, while training without stop-gradient operations. Stop-gradient yields $std$ value near $\frac{1}{\sqrt{d}}$ across each channel for all samples preventing such behaviour~\cite{simsiam}. Second, it is useful when we have only positive discriminative cases. SLOF does not consider radial distortion mitigation via changing/transforming the convolution layers but rather learns the equivariant properties of 360 videos via siamese representation. We claim that such transformation is trivial, based on the following fact. First, the omnidirectional videos are projected in angular domain, \textit{w.r.t.} $\textbf{polar}(\theta),\, \textbf{azimuthal}(\phi); \theta\in(-\frac{\pi}{2}, \frac{\pi}{2}), \phi\in(-\pi, \pi)$, so we can learn flow fields in these domains and convert these flow fields to the spherical domain using planar to spherical transformations as shown in Eq.~\eqref{eq:planetosphere} and Eq.~\eqref{eq:spheretoplane}. Second, the intent of a convolution operator in optical flow architecture is relatively different from other applications like classification, detection, or segmentation network, where other tasks require convolution to learn relevant features (spatially consistent), the relevance of these features should stay consistent (strictly for better performance) throughout any spatial location of the images/videos. However, the convolution operation is dedicated to computing the pixel-wise displacement regardless of spatial inconsistency in the distorted region via equivariant representation learning~\cite{simsiam}. Another important consideration of such a design is to make this method portable to any existing optical flow architecture. This design will eliminate the cumbersome architecture re-adjustments tasks and make it powerful and portable.

\noindent \textbf{Mapping Flow Field to Unit Sphere.} Input to our model are equirectangular images projected in angular domain $\textbf{polar}(\theta),\textbf{azimuthal}(\phi)$, where these angles are defined in radian as $\theta\in(-\frac{\pi}{2}, \frac{\pi}{2}),\phi\in{(-\pi, \pi)}$, thus the predicted optical flow is in $(\theta, \phi)$. These flow fields can be converted to a unit sphere using planar to spherical coordinate transformation as shown below:
\begin{equation}
  (x_s,y_s,z_s) = (\sin{\theta}\cos{\phi}, \sin{\theta}\sin{\phi},\cos{\theta}).
  \label{eq:planetosphere}
\end{equation}
We can compute sphere to catadioptric plane~\cite{catdiop} projections to express the flow field in Cartesian coordinates as:
\begin{equation}
 (x,y) = (\frac{x_s}{1-z_s},\frac{y_s}{1-z_s}) = (\cot{\frac{\theta}{2}\cos{\phi}}, \cot{\frac{\theta}{2}\sin{\phi}}).
  \label{eq:spheretoplane}
\end{equation}

\noindent\textbf{Design of SLOF architecture.} Given a pair of input image sequence $X_1=(x_1,x_2)$, the rotation head $(R)$ computes augmented view of this sequence as $X_2=(x^{\prime}_1, x^{\prime}_2)$ with rotation $r$ using a random combination of ``pitch", ``yaw" and ``roll" operations. These two augmented views are passed as an input to an encoder network $f$, defined as $f={P(R^\prime(\Theta(E(R(X, r)))))}$ where $E$ is a flow prediction module, RAFT~\cite{raft} in our case, $\Theta$ is a mapping of 2D flow to the unit sphere, $R^\prime$ is a reverse rotation operation and $P$ is a convolution-based down-sampling head. A prediction head presented as $h$ (an MLP head), transforms the output from the encoder $f$ from one stream to match the other stream. The illustration of this process shown in Eq.~\eqref{eq:cosine} as maximization of cosine similarity two views from the siamese stream:
\begin{equation}
  D(p^{\text{left}}, z^{\text{right}}) =- \frac{p^{\text{left}}}{||p^{\text{left}}||_2} \cdot \frac{z^{\text{right}}}{||z^{\text{right}}||_2}.
  \label{eq:cosine}
\end{equation}
Here, $p^{\text{left}}\triangleq h(f^{\text{left}}(X_1))$ and $z^{\text{right}}\triangleq f^{\text{right}}(X_2)$ denotes the output vectors to match from two different streams$(f^{\text{left}},f^{\text{right}})$. This maximization problem can be viewed from another direction, with $(p^{\text{right}}, z^{\text{left}})$ as the second matching pair from siamese stream $(f^{\text{right}},f^{\text{left}})$ respectively. Given two matching pairs, we can use the following Eq. \ref{eq:simloss} symmetrized similarity loss function (note that $z^{\text{left}}$ and $z^{\text{right}}$ are treated as a constant term using stop-grad operations to prevent a degenerate solution due to model collapse~\cite{simsiam}):
\begin{equation}
  \mathcal{L}_\text{sim} =\frac{1}{2}D(p^{\text{left}},z^{\text{right}})+\frac{1}{2}D(p^{\text{right}},z^{\text{left}}).
  \label{eq:simloss}
\end{equation}
Similarly, the optical flow loss is computed as a sequence loss~\cite{raft} over the predicted flow field and ground truth. This loss ($l_1$ distance over predicted and ground truth flow $f_\text{gt}$) is computed and averaged over a sequence of predictions iteratively generated for the same pair of input frames $\{f_1, f_2, ..., f_n\}=E(R(X,r))$ as shown in Eq.~\eqref{eq:epeloss}, where $\gamma=0.8^{n - i - 1}$ served as weights over sequence loss. Note that $(n,i)$ denotes the number of prediction$(n)$ in sequence and prediction id$(i)$ in predicted flow sequences. The design of the weighted schemes ensures different levels of confidence on predicted flows over time.
\begin{equation}
  \mathcal{L}_\text{flow} =\sum_{i=1}^{n}{\gamma||R(f_\text{gt},r) - f_i||}.
  \label{eq:epeloss}
\end{equation}
Given similarity loss$(\mathcal{L}_\text{sim})$ and flow loss$(\mathcal{L}_\text{flow})$ we implement a hybrid loss function $\mathcal{L}=\mathcal{L}_\text{sim}+\mathcal{L}_\text{flow}$. The overall objective of this loss function is to maximize the similarity between the latent representation of flow information while minimizing the loss between ground truth and predicted optical flow.

\begin{figure*}[!t]
  \centering
  \includegraphics[width=0.85\linewidth]{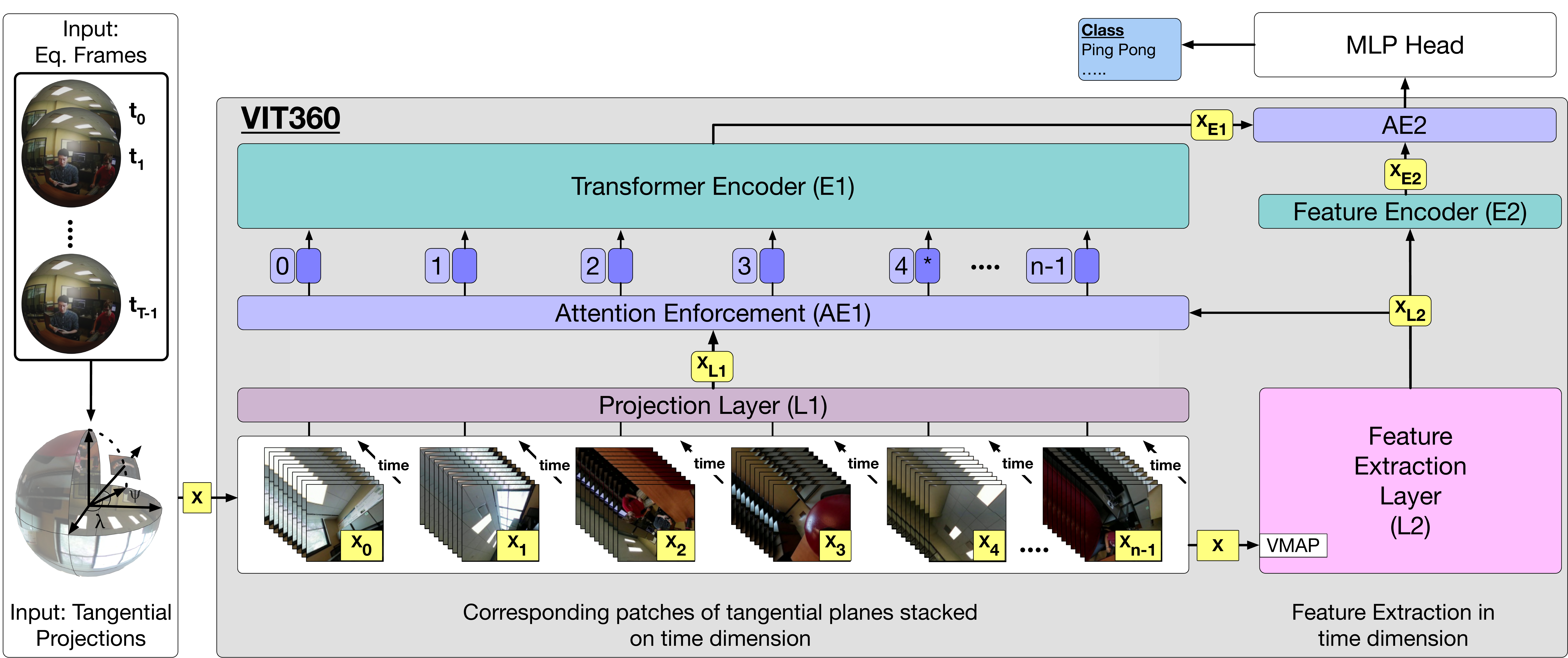}  
  \caption{\textbf{VIT360 Framework.} The sequence of ten consecutive frames (or optical flow) is sampled into six different tangential projections. These tangential projections are passed through Projection Layer (L1) and Feature Extraction Layer (L2) to compute features $X_\text{L1}$ and $X_\text{L2}$. The First Attention Enforcement Layer (AE1) computed the attention map between these features and passed it to Transformer Encoder (E1) with positional embedding as an input. The final Attention Enforcement Layer (AE2) computes the attention maps between the output $X_\text{E1}$ and $X_\text{E2}$ and passes it as an input to MLP Head for activity classification.}
\label{fig:ego_arch}
\end{figure*}
\begin{figure}[t]
  \centering
  \includegraphics[width=\linewidth]{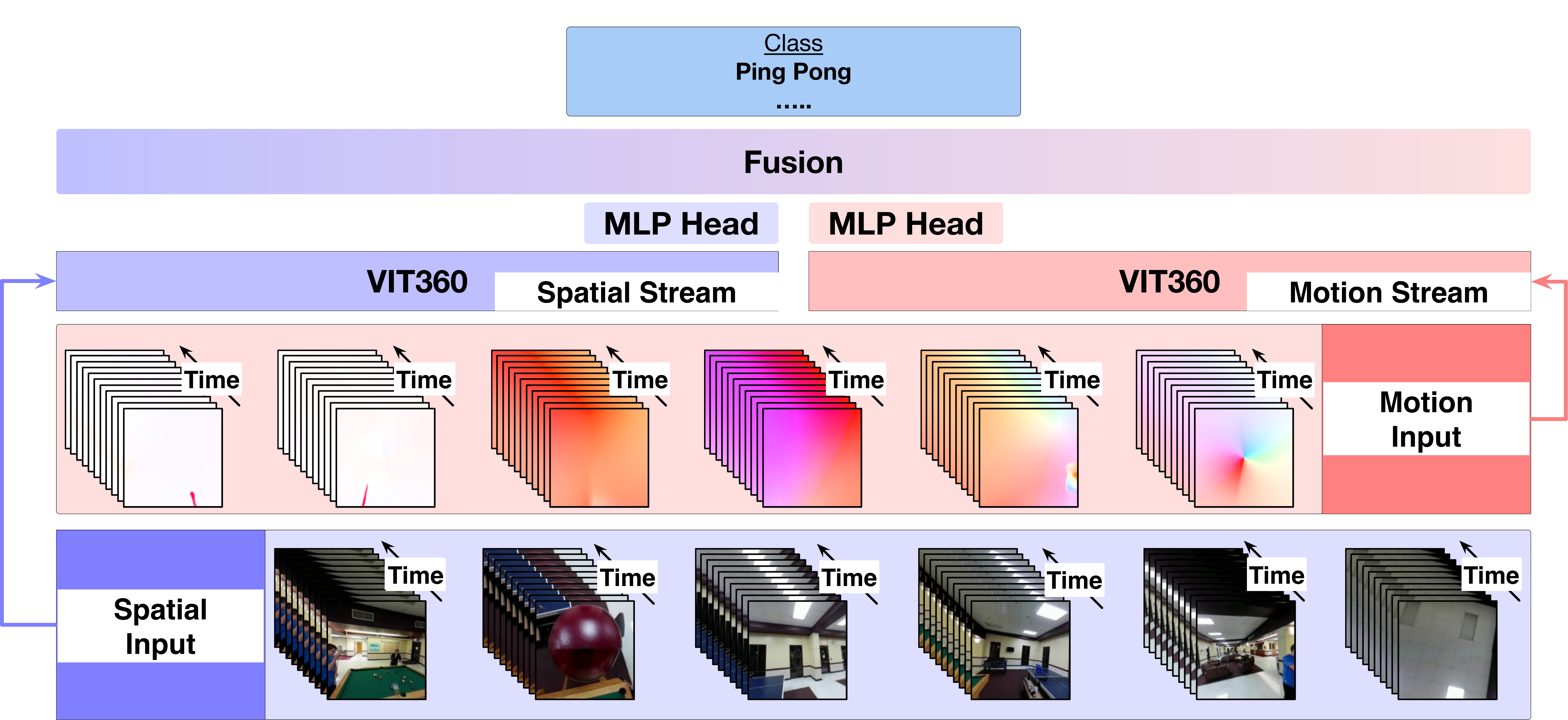}
  \caption{\textbf{Two-stream Architecture. } The motion and spatial stream constitute the component of two-stream architecture. The spatial stream receives RGB frames, whereas the motion streams receive the pre-computed optical flow. These two streams are trained independently and later fused to incorporate two different modalities via average and concatenation-based late fusion techniques. 
  }
\label{fig:classifier}
\end{figure}

\section{360$^\circ$ Egocentric Activity Recognition}
\label{sec:ego_experiment}
\begin{figure*}[t]
  \centering
  \includegraphics[width=\linewidth]{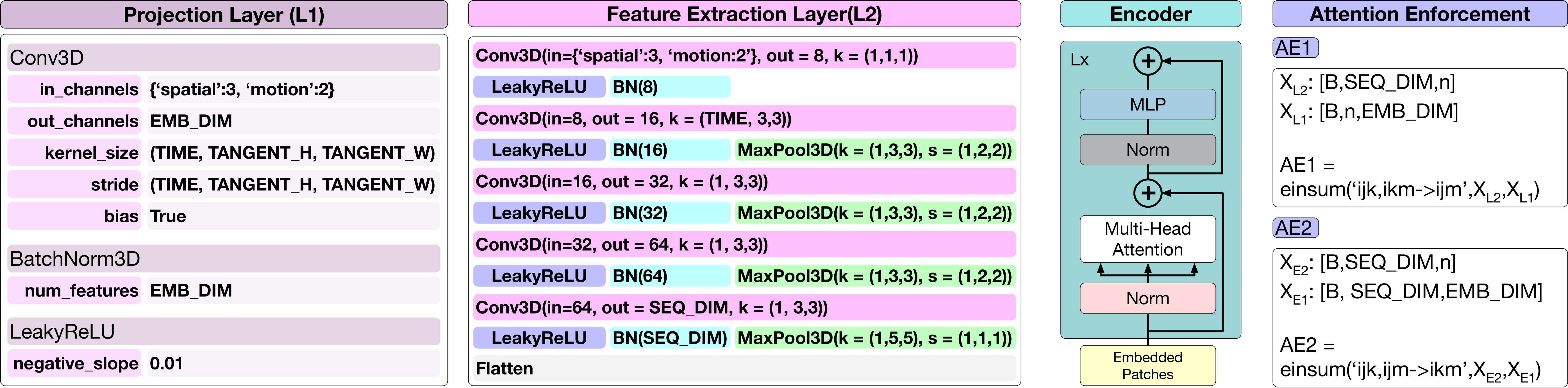}
  \caption{\textbf{Layers in VIT360. } VIT360 is composed of four major components: Projection Layer (L1) computes the initial flattened features of input sequences, and Feature Extraction Layer (L2) computes additional features using the siamese stream. Attention Enforcement Layers (AE1 and AE2) compute the attention maps computed across two different feature extraction layers and encoder streams, respectively. Finally, a pair of Transformer Encoders (E1 and E2) to learn temporal features for activity recognition.  }
\label{fig:ego_layers}
\end{figure*}
\subsection{VIT360 for Egocentric Activity Recognition}
The overall design of egocentric activity recognition in 360$^\circ$ videos comprises four different components: (1) design of VIT360, (2) two-stream approach for activity classification, (3) learning projection invariant representation of 360$^\circ$ videos, and (4) computing 360$^\circ$ or omnidirectional aware motion features from off-the-shelf architecture. The following subsections will discuss each of these components in chronological order.

\noindent \textbf{Design of VIT360.}
VIT360 requires input pre-processing which converts an equirectangular image into multiple tangential patches covering the entire FOV as shown in Fig.~\ref{fig:ego_arch}. Instead of processing raw image patches in the time dimension, these patches are processed first via the global projection layer (L1) to compute feature sequences. Compared with the original VIT~\cite{vit} implementation, input features to VIT360 are relatively larger as the patches are fed in the time dimension. The projection layers transform the larger input feature space into a relatively smaller feature space by downsizing the transformed feature dimension. The feature extraction layer (L2) acts as a siamese stream for learning additional features from independent tangent patches in the time dimension to mitigate the low parametrization of features. Using a projection layer (L1) and feature extraction layer (L2) with stacked convolution architecture makes VIT360 a hybrid architecture.

\noindent \textbf{Two-stream Architecture.}
Our egocentric activity recognition framework (shown in Fig.~\ref{fig:classifier}) is based on recent success on action/activity recognition task~\cite{twostream} loosely based on two stream hypothesis~\cite{twostreamhypothesis}. One stream is designed as a spatial stream responsible for object recognition, and another is designed as a motion stream responsible for motion recognition. Both streams of this framework implement VIT360, as discussed above, for different input modalities (e.g. frames for spatial and optical flow for motion). These streams are trained independently and later fused (late fusion) to make a joint prediction based on spatial and motion information. Following the best practices in previous work~\cite{egok360}, we implement two different fusion techniques - average and concatenation. The average fusion technique computes the average of the scores from the last layer and computes the model confidence, and does not require additional training. The concatenation-based techniques require concatenation of the last MLP head and the introduction of additional fully connected layers matching the output class dimension. This technique requires additional fine-tuning and achieves marginally better results. 

\noindent \textbf{Tangential Projections:} VIT360 takes input of 360$^\circ$ video frames or optical flows in time dimension $(X\in\mathbb{R}^{C \times T \times H\times W})$, resulting in spatial modality (where C = 3) and motion modality (where C = 2) respectively. Note that $T$ represents the number of consecutive frames denoting the time dimension. Each input at time $t$ is in fact projected in the equirectangular plane$(X^{t}\in\mathbb{R}^{C \times H\times W}, 0 \leq t\leq T)$, which is obtained by a sphere mesh unwrapped on a flat rectangular plane surface. This process maps sphere latitude and longitude to horizontal and vertical coordinate systems expressing the length and height of the plane in the range $(-\pi, \pi)$ and $(-\pi/2, \pi/2)$ respectively. In order to create input patches, tangential planes are sampled using a spherical-to-cartesian coordinate transformation system. This is shown in Eq.~\eqref{cart1}, where $(\lambda, \psi)$ represents latitude and longitude such that  $(\lambda_0, \psi_0) = (0,0)$ represents the centre of the plane, and $(c)$ represents the angular distance of the $\text{point}(x,y)$ from the centre of the projection where
\begin{equation} 
\label{cart1}
\begin{split}
    x &= \frac{\cos{\psi} \sin(\lambda - \lambda_0)}{\cos(c)},\\
    y &= \frac{\cos{\psi_0}\sin{\psi} - \sin{\psi_0}\cos{\psi}\cos{(\lambda - \lambda_0)}}{\cos{c}},\\
    \cos(c) &= \sin{\psi_0}\sin{\psi} + \cos{\psi_0}\cos(\psi)\cos(\lambda - \lambda_0).
\end{split}
\end{equation}
Similarly, the inverse map from plane to the sphere can be computed using the following equation as
\begin{equation}
\label{cart2}
\begin{split}
    \psi &= \sin^{-1}{\bigg(\cos{(c)}\sin{\psi_0} + \frac{y\sin(c)\cos{\psi_0}}{\rho}\bigg)},\\
    \lambda &= \lambda_0 + \tan^{-1}{\bigg(\frac{x\sin{(c)}}{\rho\cos{\psi_0}\cos{(c)} - y\sin{\psi_0}\sin{(c)}}\bigg)},\\
    \quad \text{s.t.}\quad \rho &= \sqrt{x^2+y^2},~ c = \tan^{-1}{\rho}.
\end{split}
\end{equation}
The projection layer samples $n$ tangential planes $(\mathbb{R}^{C \times T \times h \times w})$ covering the entire 360$^\circ$ FOV, where $(h,w)$ are the height and width of each tangential plane. These tangential planes are linearly stacked alongside the width to obtain a final output $x\in\mathbb{R}^{C \times T \times h \times nw}$. In summary, the tangential projection layers take a series of optical flows or frames in a video $(X\in\mathbb{R}^{C \times T \times H \times W})$ and transform it into a series of tangential plane projections $(x\in\mathbb{R}^{C \times T \times h \times nw})$ linearly stacked along the width.


\begin{figure*}[t]
  \centering
  \includegraphics[width=0.8\linewidth]{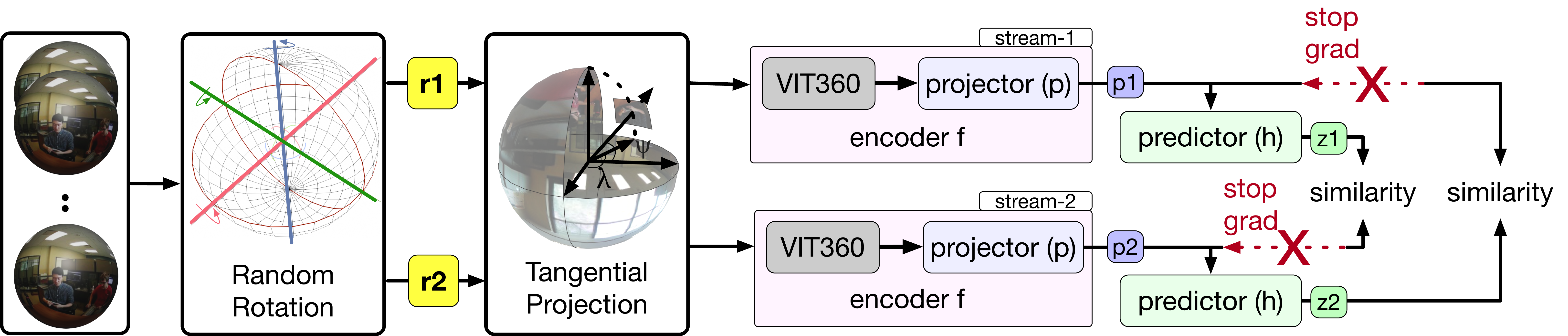}
  \caption{\textbf{Siamese Representation Learning for egocentric activity recognition.} The random Rotation head (R) computes two different rotational views of the same input video sequences. These augmented views are passed as input to VIT360 using tangential projections. The siamese network of VIT360 computes latent representation, $p_i$ and $z_i$, using projector$(p)$ and predictor $(h)$ respectively. The training policy maximizes the similarity between latent representations $(p_1, z_2)$ and $(p_2, z_1)$ in an unsupervised manner. The pre-trained network with this policy is then subjected to training for egocentric activity recognition.}
\label{fig:ego_representation}
\end{figure*}
\noindent \textbf{Projection Layer (L1):} An overview of the Projection Layer (L1) is shown in Fig.~\ref{fig:ego_layers}. This layer takes tangential patches~$(x\in\mathbb{R}^{C \times T \times h \times nw})$ in time dimension as input. In contrast with the original VIT implementation, the input to VIT360 is a  3D image (considering the time dimension) which requires a modification of the original projection layer to 3D convolution-based architecture. The Projection Layer (L1) includes a 3D convolution with 3D batch normalization and LeakyReLU layer as an activation function. The 3D convolution layer is parametrized with the 3D kernel size and stride as (TIME, TANGENT\_H, TANGENT\_W), where TANGENT\_H = h, TANGENT\_W = w refers to the height and width of tangent patches and TIME = T refers to the number of frames in the time dimension. This formulation of L1 transformed input tangential patches $(x\in\mathbb{R}^{C \times T \times h \times nw})$ to $(x\in\mathbb{R}^{n \times \text{EMBED\_DIM}})$, where \text{EMBED\_DIM} refers to the output feature dimension in Fig.~\ref{fig:ego_layers}.

\noindent \textbf{Feature Extraction Layer (L2): } The Projection Layer (L1) inspired by original VIT implementation maps a huge feature space $(x\in\mathbb{R}^{C \times T \times h \times nw})$ into relatively smaller feature $(x\in\mathbb{R}^{n \times \text{EMBED\_DIM}})$ embedding. This low parametrization of the feature space creates VIT360 to suffer from significant performance degradation. To mitigate this issue, an additional Feature Extraction Layer (L2), as shown in Fig.~\ref{fig:ego_layers} is designed to learn additional features. These additional features are later used for attention enforcement, resulting in improved performance. The Feature Extraction Layer (L2) is implemented as a siamese network to process individual input tangential patches. To achieve the siamese network, we use \textbf{vmap} (available in PyTorch framework) operation per tangential patches. The Feature Extraction Layer (L2) takes an input $(x\in\mathbb{R}^{C \times T \times h \times nw})$, reshapes it to $(x\in\mathbb{R}^{n \times C \times T \times h \times w})$ and computes features $(x\in\mathbb{R}^{n \times \rm{SEQ\_DIM}})$ where the \rm{SEQ\_DIM} is an output embeddings of the Feature Extraction Layer (L2). 

\noindent \textbf{Attention Enforcement (AE1 and AE2): } The Encoder in the original implementation of VIT takes input from the linear projection layer from $16\times16$ image patches. Such design consideration in VIT360 is computationally challenging since the input for VIT360 is stacked patches in the time dimension. It is possible to generate a relatively larger number of tangential planes and subsequently larger embedding space from the Projection Layer (L1). Such a design is computationally infeasible because the input will be relatively larger than the original VIT. This tradeoff between the choice of the number of input patches and computational cost directly affects model performance. In addition, the Projection Layer (L1) reduces the feature space significantly, as we discussed above, leading to lower parametrization and reducing the model's performance. In order to mitigate such issues and maintain the optimal input size, we introduce Attention Enforcement (AE1 and AE2) techniques. The first Attention Enforcement (AE1) computes an attention map between the output from L1$(X_\text{L1})$ and L2$(X_\text{L2})$ making the input $(X\in\mathbb{R}^{\rm{SEQ\_DIM} \times \rm{EMB\_DIM}})$ size of the encoder sufficiently optimal retaining the model performance. Similarly, the second Attention Enforcement(AE2) computes the attention maps $(X\in\mathbb{R}^{n \times \rm{SEQ\_DIM}})$ between the output from E1$(X_\text{E1}\in\mathbb{R}^{\rm{SEQ\_DIM} \times \rm{EMB\_DIM}})$ and E2$(X_\text{E2}\in\mathbb{R}^{\rm{SEQ\_DIM} \times n})$, keeping the number of parameter~$(n \times \rm{EMB\_DIM} \times NUM\_CLASSES)$ in MLP Head fixed. The overview of Attention Enforcement is presented in Fig.~\ref{fig:ego_layers}.

\noindent \textbf{Encoder Layer (E1 and E2): } The Transformer Encoder layer E1 and E2 contains similar implementation as presented in the original VIT implementation. The E1-encoder contains two layers of encoder architecture, whereas E2 contains only one layer of encoder architecture. E1 takes an input~$(X_{AE1}\in\mathbb{R}^{\rm{SEQ\_DIM} \times \rm{EMB\_DIM}})$  from the Attention Enforcement Layer (E1) where \text{EMBED\_DIM} is the input dimension of the E1-encoder. Similarly, E2 takes an input~$(X_\text{L2}\in\mathbb{R}^{n \times \rm{SEQ\_DIM}})$  from the Feature Extraction Layer (L2) where \rm{SEQ\_DIM} is the input dimension of the E2-encoder. Both E1 and E2 computes output feature~$(X_\text{E1}\in\mathbb{R}^{\rm{SEQ\_DIM} \times \rm{EMB\_DIM}}, X_\text{E2}\in\mathbb{R}^{n \times \rm{SEQ\_DIM}})$ vectors similar to the input dimension.

\noindent \textbf{MLP Head: } In contrast to the original VIT implementation, we introduce a single-layer MLP head as a classification head to predict the activity classes. The Attention Enforcement Layer (AE2) output is passed as an input to the final MLP Head.

\noindent \textbf{Projection Invariant Representation:}
360$^\circ$ videos in comparison with perspective videos exhibit a unique property of unlimited FOV, resulting in infinite projections. Such infinite projections can be achieved by rotating the 360$^\circ$ videos on three different axes $(X, Y, Z)$, namely ``pitch'', ``roll'', and ``yaw'' operations. Regardless of these projections, the overall information of 360$^\circ$ videos remains intact. This rotation-invariant property of 360$^\circ$ videos is crucial while designing deep learning architecture for 360$^\circ$ videos based applications. In practice, the ideal architectures should be rotationally invariant or projection invariant. To achieve this goal, we perform pretraining of VIT360 to learn projection invariant representation.

\begin{figure*}[t]
  \centering
  \includegraphics[width=0.85\linewidth]{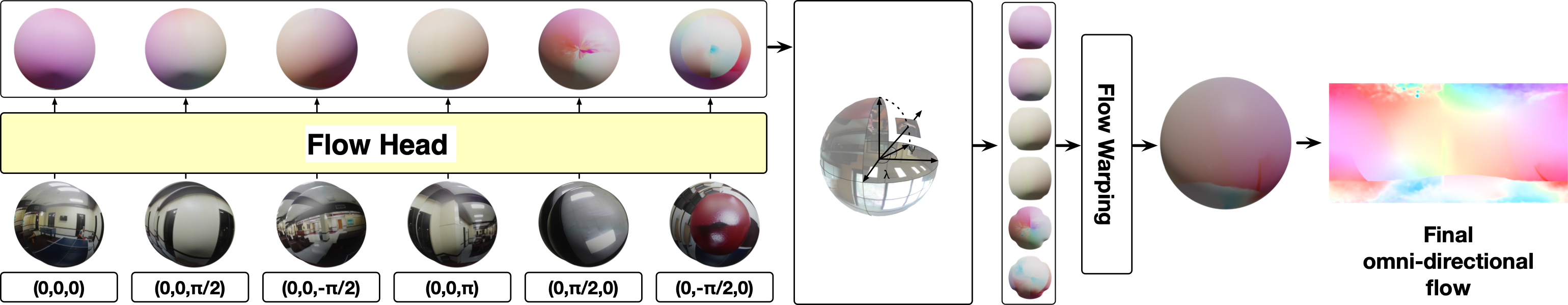}
  \caption{\textbf{360$^\circ$ flow inference using PercieverIO.} The pairs of consecutive frames are projected with six different FOV covering the entire 360$^\circ$ field. The PercieverIO~\cite{perceiverio} is implemented as Flow Head, which computes the optical flow in each pair of projected FOV. The computed flow fields are then projected using tangential projections to extract the central FOV. This central FOV is then warped in a spherical representation and the final omnidirectional flow is computed using sphere-to-plane projections of the warped spherical flow. }
\label{fig:floarch}
\end{figure*}

Pre-training of VIT360 for projection invariant representation is loosely based on a contrastive learning approach called SimSiam~\cite{simsiam, bhandari2022learning}. The overview of pre-training stage is illustrated in Fig.~\ref{fig:ego_representation}. This representation learning based on SimSiam is completed using siamese representation learning, formulated as a pair of weight-sharing VIT360 streams (stream-1, stream-2) to maximize the similarity between two different representations of the same 360$^\circ$ videos information (both frames and optical flow). Given a sequence of frames $X = (x_0, x_1, ..., x_{T-1})$, where $T$ is the number of frames in sequence in time, the rotation head $(R)$ computes a pair of rotational augmentation $X_i = R(X,r_i)$ using $r \in (r_1,r_2)$ defined as a random configuration of ``pitch'', ``yaw'' and ``roll'' operations. These rotationally augmented representations are now passed as an input to an encoder network $f=\textbf{h}(\textbf{p}(\text{VIT360}(R(X,r))))$, where $\textbf{p}$ and $\textbf{h}$ are MLP layers, defined as the projector and predictor's head respectively. A projector head $(\textbf{p})$ appends the MLP Head (shown in Fig.~\ref{fig:architecture}) with additional MLP layers to create a bottleneck MLP module, which constitutes the final VIT360 architecture. Similarly, a predictor head $(\textbf{h})$ transforms the output $(p_i\in\mathbb{R}^{n\times \text{EMBED\_DIM}})$ from the encoder $(f)$  to $(z_i\in\mathbb{R}^{n\times \text{EMBED\_DIM}})$ where $i = {1,2}$ represents the first and second stream. 

Given output $(p_i,z_i)$ from respective streams, we formulate a pre-training stage to maximize cosine similarity between these representations across siamese streams. The illustration of the maximization process is shown in Eq.~\eqref{cosine}.
\begin{equation} 
\label{cosine}
\begin{split}
    D(p_1, z_2) & = -\frac{p_1}{||p_1||_2}\cdot\frac{z_2}{||z_2||_2}, \\ \mathcal{L}_\text{sim} &= \frac{1}{2}D(p_1, z_2) + \frac{1}{2}D(p_2,z_1).
\end{split}
\end{equation}
Here, $p_1=f(X_1)$ and $z_2 = h(f(X_2))$ denote latent representations computed at different levels for the same input $X$ using rotationally augmented different view $(X_1,X_2)$. Siamese VIT360 is trained so that both projector and predictor layers are subjected to learn a similar latent representation of the given input across two different sets of inputs.  This representation learning aims to maximize the similarity between these two outputs at different levels across the siamese streams. The maximization problem can be considered in both directions, another being the maximization between $(p_2, z_1)$. Given output pairs $(p_1, z_2)$ and $(p_2, z_1)$, we use the symmetric similarity loss function $(\mathcal{L}_\text{sim})$ as shown in Eq.~\eqref{cosine}. As discussed in SimSiam~\cite{simsiam}, we treat $(p_1,p_2)$ as a constant via stop-grad operations to prevent a degenerate solution due to model collapse.

\subsection{Omnidirectional-aware Optical Flow Estimation}
The overview of the optical flow inference for 360$^\circ$ videos using PerceiverIO is shown in Fig.~\ref{fig:floarch}. Similar to VIT360, this approach also considers different projections of pairs of input 360$^\circ$ frames for calculating the final flow between the frames. These different projections are considered an input patch to PercieverIO, which it calculates motion information in each pair. Later these motions in patches are warped into one 360$^\circ$ FOV to compute the final 360$^\circ$ optical flow. In order to maintain the computational size, we limit the number of patches to six and perform six different projections to cover the entire 360$^\circ$ FOV. Compared to the original PercieverIO implementation, we do not crop these different projections but rather resize these projections into input size as defined in PercieverIO. We do not limit the flow computation across the neighboring patches by avoiding cropping beyond the padding zone. This intermediate flow represents optical flow per projections over the entire equirectangular plane. Given optical flow computed in an equirectangular plane, flow information in the highly distorted area (like polar regions) suffers the impact of distortions. However, the central part of equirectangular plane $(\lambda_0, \psi_0) = (0,0)$ is the least distorted region from where the tangential patches are sampled with $\frac{\pi}{2}$ FOV. These sampled patches cover the entire 360$^\circ$ FOV. Finally, these patches are warped into global 360$^\circ$ flow for input frames.
\begin{figure}[!t]
\centering
    \includegraphics[width=0.9\linewidth]{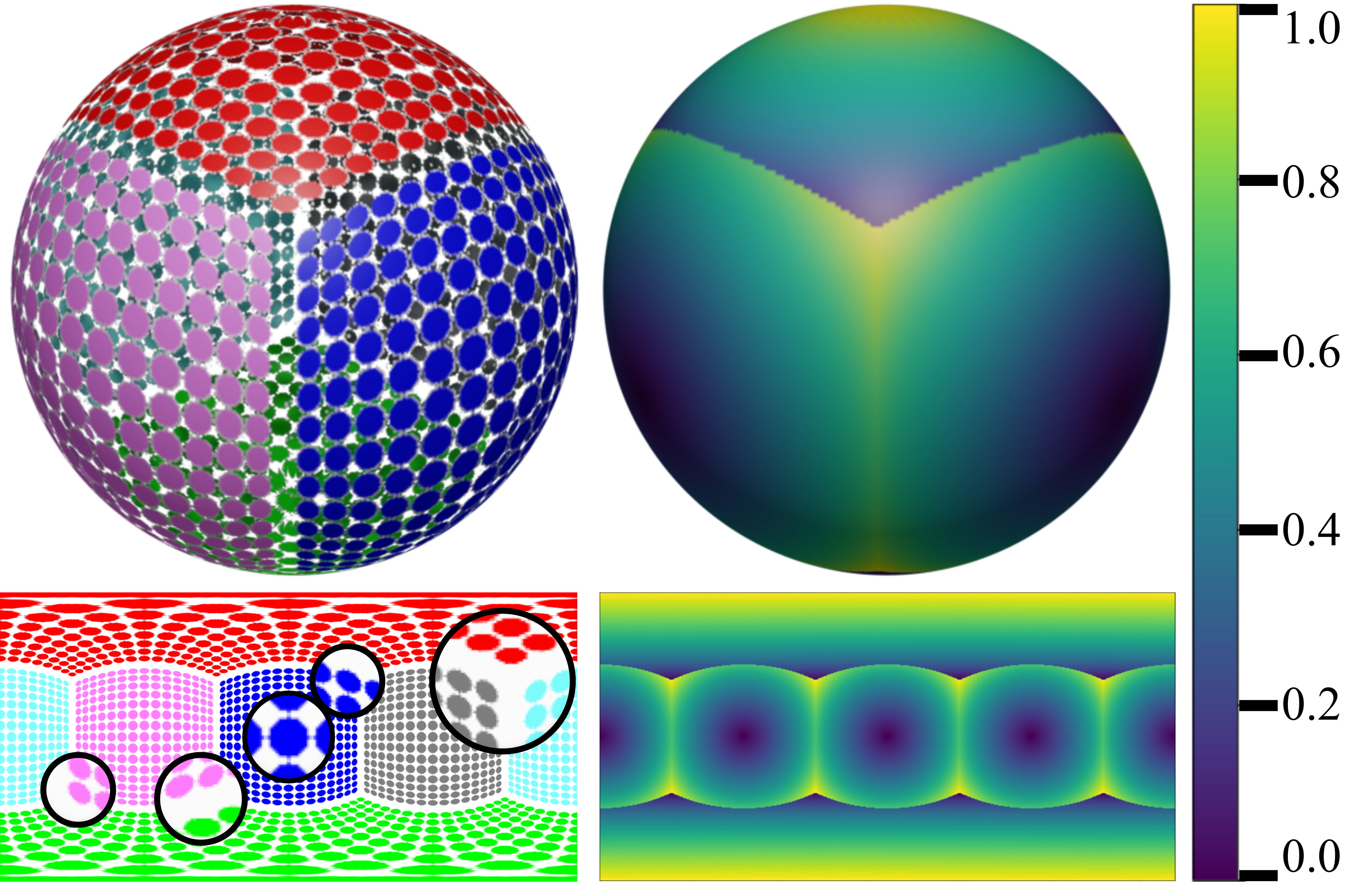}
    \caption{\textbf{Distortion density map}.  Illustrating different distortion intensities due to equirectangular projections. Left: upper (\textcolor{red}{red}) and lower (\textcolor{green}{green}) part of projections shows higher distortion in the central part whereas the equatorial region (\textcolor{cyan1}{cyan}, \textcolor{pink1}{pink}, \textcolor{blue}{blue}, \textcolor{gray}{gray}) exhibit higher distortion rate away from the center of the tangential plane. Right: shows the distortion density from $(0,1)$. This distortion density map is used to evaluate the distortion-aware EPE (EPE$_d$). \textbf{Note}: Each circle patch in the left spherical projection has the same area.}
    \label{fig:distortiondensity}
\end{figure}
\begin{figure*}[!t]
    \centering
    \includegraphics[width=\linewidth]{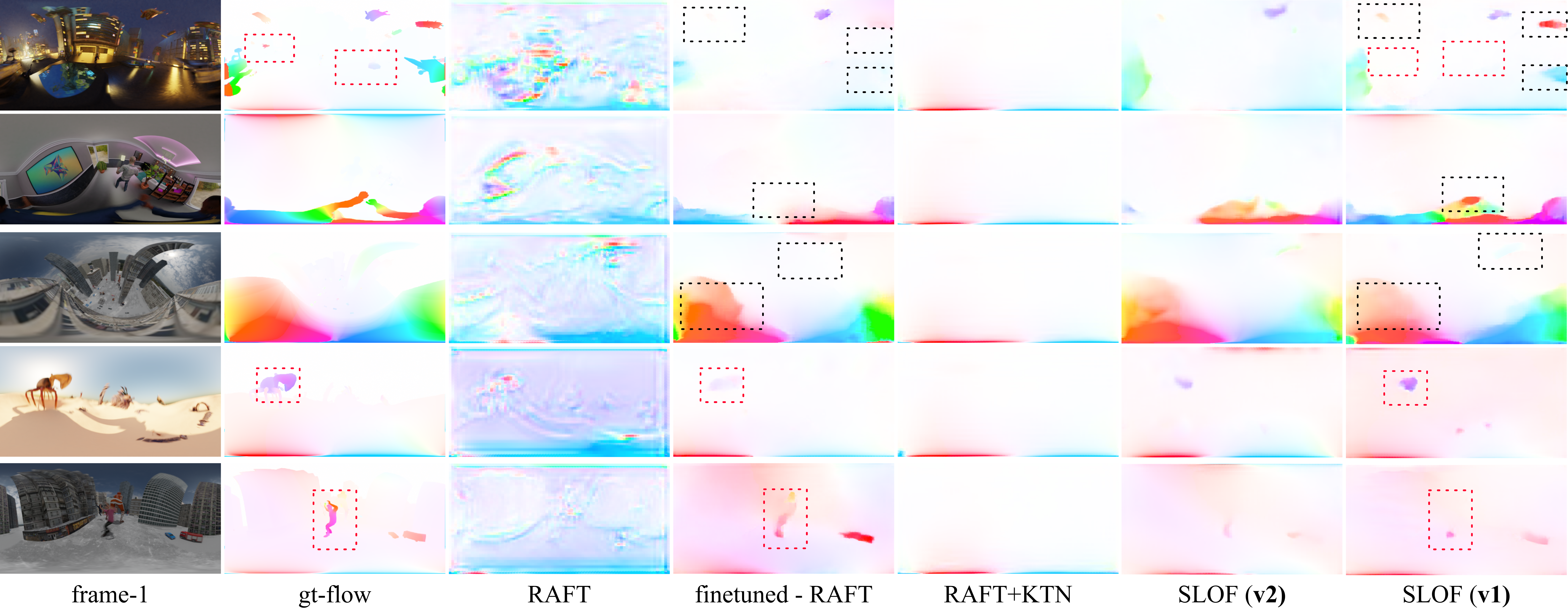}
    \caption{\textbf{Qualitative results on FLOW360 test set}.  Qualitative results show our best model SLOF(\textbf{v1}) shows better results compared to fine-tuned RAFT trained with policy explained in~\cite{raft}. The dotted (\textbf{black}) rectangle indicates the comparative improvements of our model over fine-tuned RAFT. RAFT+KTN method fails to predict flow-field correctly; instead, it only predicts shallow flow fields from camera motion. The weakness of our model can be seen on the dotted (\textcolor{red}{red}) rectangle where smaller motion segments are missing. \textbf{Note:} Flows information is clipped for better visualization.}
    \label{fig:qualitative}
\end{figure*}
\section{Experiments}
\subsection{360$^\circ$ Optical FLow Estimation}
We evaluate SLOF on the FLOW360 test set. We use pre-trained RAFT on Sintel~\cite{sintel} and fine-tune on FLOW360 as a comparison baseline. The fine-tuning process is done using training protocols suggested in \cite{raft}. Moreover, to make a fair comparison with traditional methods, we transform RAFT (pre-trained) to adapt spherical convolution using KTN~\cite{ktn}. KTN transforms the convolution kernel to mitigate the radial distortions via estimating the spherical convolution function. Additionally, we run ablation studies on different training strategies and propose a distortion-aware evaluation. We will present details of the training procedure in the supplemental material.
\begin{figure*}[!t]
\centering
    \includegraphics[width=0.96\linewidth]{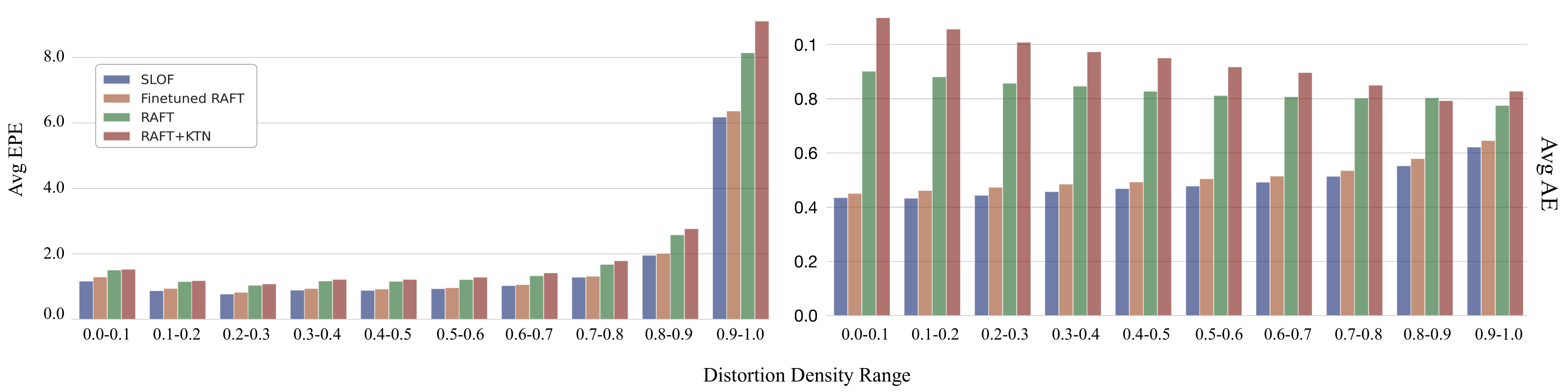}
    \caption{\textbf{Error distribution plot of EPE and AE in different distortion density ranges.} SLOF performs better in all distortion density ranges.}
    \label{fig:histoplot}
\end{figure*}
\begin{table*}[!t]\small
\centering
\caption{\textbf{Quantitave results on FLOW360 test set.} $\ast$ denotes that we use EPE$_d$/AE$_d$ as the metrics; otherwise, the normal EPE and AE. Compared to the baseline, SLOF achieves lower end-point error and angular error on both distortion aware (EPE$_d$ and AE$_d$) and normal scheme. In terms of end-point error (lower the better) our model (\textbf{v1}, \textbf{v2}) outperforms all the baselines. Similarly, in terms of angular error (lower the better) our models (\textbf{v1}, \textbf{v2}) perform comparatively similarly and outperform all the baselines. Though RAFT+KTN achieves comparable normal EPE, the distortion-aware (Weighted) metrics (EPE$_d$ and AE$_d$) are significantly larger. \textbf{Note}: metrics in range (all, less than (5, 10, 20) and greater than 20) is computed as an average, based on the speed (\textbf{s}(x,y)$=\sqrt{u(x,y)^2+v(x,y)^2}$) only in the respective pixel regions.}
\scalebox{0.9}{
\begin{tabular}{c|c|ccccccccc}
\toprule
\rowcolor{white}
Method & Version & Metric & Weighted \textbf{s}$\geq$0$^\ast$ & \textbf{s}$\geq$0 & \textbf{s}$<$5 & \textbf{s}$<$10 & \textbf{s}$<$20 & \textbf{s}$\geq$20 \\
\hline
\multirow{6}{*}{Baselines} & \multirow{2}{*}{RAFT~\cite{raft}} & EPE & 3.344 & 2.058 & 0.558 & 0.682 & 0.838 & 71.736 \\
& & \cellcolor{Gray}AE & \cellcolor{Gray}1.120 & \cellcolor{Gray}0.820 & \cellcolor{Gray}0.825 & \cellcolor{Gray}0.821 & \cellcolor{Gray}0.819 & \cellcolor{Gray}0.868  \\
\hhline{~|-|---------}
& \multirow{2}{*}{Finetuned RAFT~\cite{raft}} & EPE & 2.635 & 1.624 & 0.314 & 0.393 & 0.509 & 65.340\\
& & \cellcolor{Gray}AE & \cellcolor{Gray}0.745 & \cellcolor{Gray}0.522 & \cellcolor{Gray}0.527 & \cellcolor{Gray}0.522 & \cellcolor{Gray}0.520 & \cellcolor{Gray}0.647  \\
\hhline{~|-|---------}
& \multirow{2}{*}{RAFT + KTN~\cite{ktn}} & EPE & 3.899 & 2.222 & 0.598 & 0.742 & 0.924 & 76.426\\
& & \cellcolor{Gray}AE & \cellcolor{Gray}2.020 & \cellcolor{Gray}0.912 & \cellcolor{Gray}0.912 & \cellcolor{Gray}0.910 & \cellcolor{Gray}0.911 &  \cellcolor{Gray}1.0114  \\
\hline
\multirow{4}{*}{SLOF} 
& \multirow{2}{*}{Switch rotation (\textbf{v2})} & EPE & 2.626 & 1.615 & 0.326 & 0.401 & 0.512 & 64.678\\
& & \cellcolor{Gray}AE & \cellcolor{Gray}\textbf{0.691} & \cellcolor{Gray}\textbf{0.485} & \cellcolor{Gray}\textbf{0.489} & \cellcolor{Gray}\textbf{0.484} & \cellcolor{Gray}\textbf{0.482} & \cellcolor{Gray}0.659  \\
\hhline{~|-|---------}
& \multirow{2}{*}{Single rotation (\textbf{v1})} & EPE & \textbf{2.548} & \textbf{1.568} & \textbf{0.309} & \textbf{0.387} & \textbf{0.502} & 62.476\\
& & \cellcolor{Gray}AE & \cellcolor{Gray}0.708 & \cellcolor{Gray}0.497 & \cellcolor{Gray}0.501 & \cellcolor{Gray}0.497 & \cellcolor{Gray}0.495 & \cellcolor{Gray}\textbf{0.607}  \\
\bottomrule
\end{tabular}}
  \label{tab:quantitative}
\end{table*}

\noindent\textbf{Scope.} The scope of our experiments is two folds: First, create a baseline for future researchers to explore novel methodologies. Second, address the validity of our method based on fair comparisons with a flow network designed for a spherical dataset. We formulate our baseline experiment on perspective optical flow network RAFT and a modified version of RAFT with KTN~\cite{ktn} to compare the performance. The RAFT+KTN architecture simulates a domain adaptation similar to approaches like \cite{revisiting,omniflownet}. We choose KTN because of its success over alternative approaches like~\cite{spectral1,spectral2,saliency,spherical2,spherical3}. It is worth noting that the design of omnidirectional flow estimation can be extended to several techniques involving the mitigation of radial distortions, making it practically impossible to cover all.

\noindent\textbf{Augmentation Strategy.} Given the nature of SLOF, we can train it using two different training strategies (\textbf{v1}, \textbf{v2}) as shown in Fig.~\ref{fig:architecture} (right). These strategies can be achieved by performing different rotational augmentation on the input sequences. The first strategy (\textbf{v1}) can be achieved by using set of inputs $(R(X_1,r_1), R(X_2,r_2))$ where $r_1=(0,0,0)$, i.e., $X_1$ does not have any rotational augmentation, whereas $r_2\neq(0,0,0)$ has rotation defined with random combinations of ``pitch", ``roll", and ``yaw" operations. This setting is kept consistent throughout the training process. Alternatively, identical augmentation can be achieved by flipping this augmentation protocols. The second rotational scheme (\textbf{v2}) can be achieved by randomly switching rotation such that when $r_1$ is none, the $r_2$ is some random rotational augmentation and vice versa. This approach performs on par with \textbf{v1}.

\begin{equation}
  \text{AE} = \arccos(\frac{u_eu_r + v_ev_r + 1}{\sqrt{u_r^2+v_r^2+1} \sqrt{u_e^2+v_e^2+1}}).
  \label{eq:ae}
\end{equation}
\begin{equation}
  \text{EPE} = \frac{1}{N}\sum_{i}^{N}{{||f_\text{pred} - f_\text{gt}||_2}}.
  \label{eq:epe}
\end{equation}
\begin{equation}
  \text{EPE}_{d} = \frac{1}{N}\sum_{i}^{N}{\frac{||f_\text{pred} - f_\text{gt}||_2}{1-d}}.
  \label{eq:distortionepe}
\end{equation}
\noindent\textbf{Evaluation Strategy.} We evaluate our method based on 2D-raw flow. Besides, using EPE (End Point Error in Eq.~\eqref{eq:epe}), i.e., Euclidean distance between the predicted flow and ground truth flow, as a single evaluation metric, we incorporate AE (Angular Error) as shown in Eq.~\eqref{eq:ae} as the second measure. To explain the error in the omnidirectional setting, we introduce a distortion-aware measure called EPE$_d$ as in Eq.~\eqref{eq:distortionepe}. This metric penalizes the error in the distorted area based on the distortion density map.

As EPE$_d$, AE$_d$ is calculated as $\frac{1}{N}\sum_{i}^{N}{\frac{\text{AE}}{1-d}}$ where, $d$ represents the distortion density map illustrated in ~Fig.~\ref{fig:distortiondensity}, $f_\text{pred}=(u_e,v_e)$ represents predicted flow, and $f_\text{gt}=(u_r,v_r)$ represents ground truth flow. Note that, to maintain lower metrics scale the distortion density is mapped between $[0.500,1.000)$ from $(0.0, 1.0]$. Please refer to supplemental for additional details on distortion density map.

\noindent\textbf{Results.} Fig.~\ref{fig:qualitative}, Fig.~\ref{fig:histoplot} and Table~\ref{tab:quantitative} summarize our experimental results. The overall summary of qualitative results is presented in Fig.~\ref{fig:qualitative}. SLOF performs better than baseline RAFT and kernel transformed RAFT+KTN methods. This result is evident enough to show that siamese representation learning can exploit the rotational properties of 360$^\circ$ videos to learn omnidirectional optical flow regardless of explicit architecture adjustments.

Our methods, SLOF (\textbf{v1}, \textbf{v2}) perform better than presented baselines. Among these methods \textbf{v1} has the best EPE score whereas, \textbf{v2} has better AE score. However, AE on both \textbf{v1} and \textbf{v2} are relatively similar, suggesting \textbf{v1} as our best method, shown in Fig.~\ref{fig:qualitative}. 

By investigating distortion-aware EPE, we can see that RAFT with KTN achieves significantly higher EPE regardless of comparable normal EPE with the other methods. This clearly explains why RAFT+KTN methods could not predict the motion around the distorted area; instead, it predicts shallow flow fields due to camera motion only. Moreover, comparing qualitative results in Fig.~\ref{fig:qualitative} and EPE measure in different distortion ranges in Fig.~\ref{fig:histoplot}, we can see that our best method can predict smoother flow fields compared to baseline methods. These fields in the polar region are comparatively better and have better motion consistency in the edge region. However, our model might fail to predict relatively smaller motion regions in some cases, which leaves room for future improvements based on the proposed method. This concludes that RAFT+KTN requires additional re-engineering and domain adaptation, which is out of the scope of current work.

\begin{table*}[!t]
\small
\caption{\textbf{Quantitave Results.} The Top-1 accuracy (shown as \%) of VIT360 compared with baseline methods in EGOK360~\cite{egok360} achieves consistent performance regardless of projection mode, achieving less than 0.40\% accuracy difference. The baseline method shows performance gaps between 5-14\%. 360$^\circ$ optical flow computed using our inference techniques boost the performance of VIT360 by almost 5\%. (Note: We use Perspective Flow (RAFT~\cite{raft}) for baseline experiments.)}
\centering
\scalebox{0.9}{
\begin{tabular}{cccccccccc}
\toprule
\rowcolor{white}
\textbf{Version} & \textbf{Random Projection} & \textbf{Flow} & \textbf{Motion} & \textbf{Spatial} & \textbf{Fused Avg.} & \textbf{Fused Concat.}\\
\hline
\multirow{2}{*}{ResNet} & \checkmark & Perspective& 42.53 & 59.67 & 56.18 & 62.79\\
&  & Perspective& 56.43 & 68.95 & 66.79 & 69.32\\
\hhline{---------}
\multirow{2}{*}{I3D} & \checkmark & Perspective & 48.43 & 63.61 & 60.17 & 65.39\\
&  & Perspective& 62.19 & 69.78 & 67.37 & 72.68\\
\hhline{----------}
\hhline{----------}
\hhline{----------}
\multirow{4}{*}{VIT360 (Ours)} & \checkmark & Perspective& 56.52 & 67.14 & 65.18 & 70.89\\
&  & Perspective& 56.57 & 67.23 & 65.23 & 70.79\\
\hhline{~---------}
\hhline{~---------}
& \checkmark & 360$^\circ$(Ours) & 60.34 & 71.29 & 70.66 & 75.87\\
&  & 360$^\circ$(Ours) & \textbf{60.43} & \textbf{71.13} & \textbf{70.29} & \textbf{75.59}\\
\bottomrule
\end{tabular}}
\label{tab:qtres}
\end{table*}
\begin{figure*}[t]
  \centering
  \includegraphics[width=0.99\linewidth]{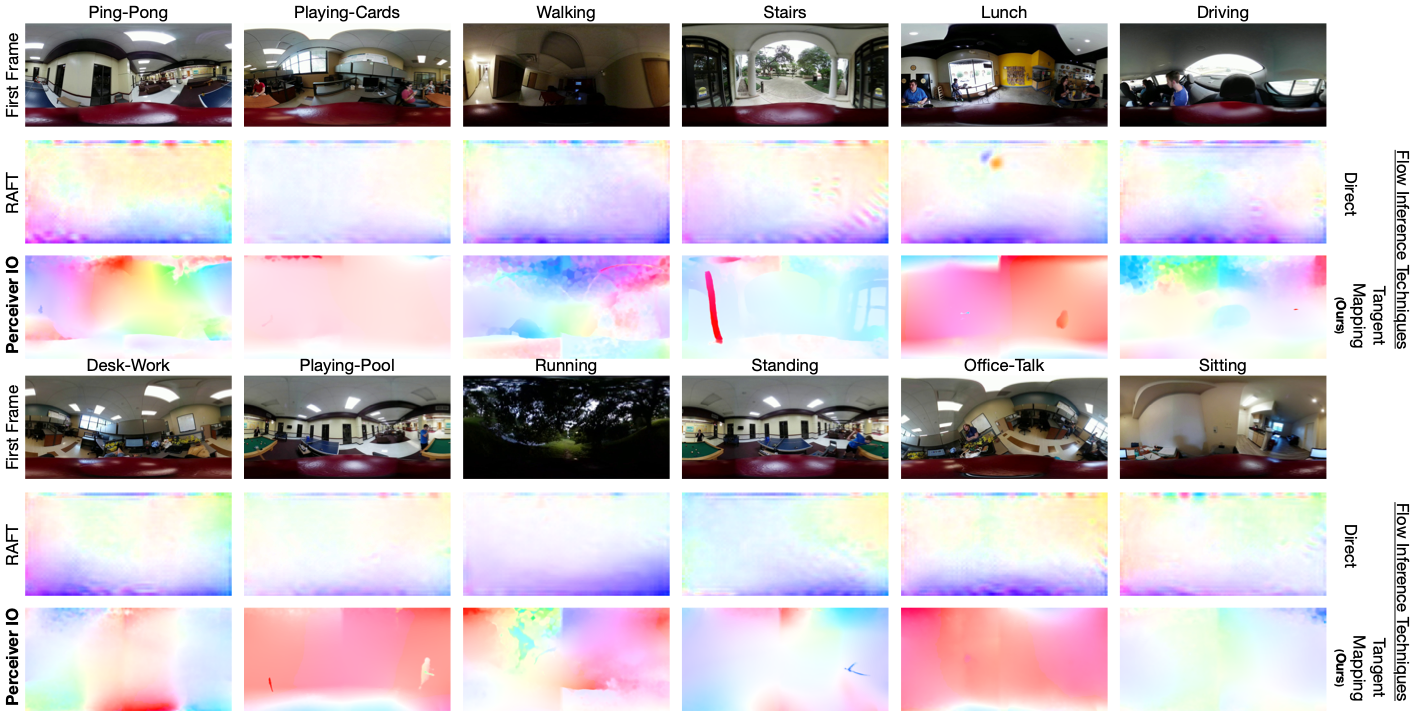}
  \caption{\textbf{360$^\circ$ optical flow results.} We compare 360$^\circ$ flow inferred using our techniques and perspective video-based technique using RAFT~\cite{raft}. The superiority of our method can be seen in optical flow with smooth motion, distinct motion region, and flow continuity across the edges.}
\label{fig:floquality}
\end{figure*}
\subsection{360$^\circ$ Egocentric Activity Recognition}
We focus our experiments mainly on representation learning based egocentric activity recognition on 360$^\circ$ videos. We run our experiments on EGOK360~\cite{egok360} dataset, an egocentric activity recognition dataset for 360$^\circ$ videos. We present a series of experiments that demonstrate the efficacy of pre-training VIT360 with a siamese representation approach maximizing the representation across the different views of the same input. In addition, we study the impact of using 360$^\circ$ or omnidirectional aware optical flow compared to traditional perspective videos based on optical flow on 360$^\circ$ videos for this motion-based task. 

The first experiment includes egocentric activity recognition (using a two-stream approach) without the representation learning scheme. Similarly, the follow-up experiments include the representation of learning-based egocentric activity recognition, which follows two different consecutive experiments of pre-training and two streams approach for classification. Similarly, we conducted two additional experiments based on optical flow and omnidirectional optical flow. In addition to these, we also make a comparative study of qualitative results on proposed PercieverIO-based 360$^\circ$ optical flow inference techniques with pre-trained perspective video-based optical flow technique, RAFT~\cite{raft}. Note that our comparisons based on the two-stream architecture are reported in EGOK360~\cite{egok360} dataset only. Since the experimental results presented on EGOK360 are based on a random train/test split of entire datasets, we also follow a similar approach. However, to make a fair test case, we sample train and test frames with a 9:1 split from each clip from the entire training data. We achieve marginally similar baseline results on EGOK360 compared to experimental results reported in the original paper. In addition, we only consider activity recognition on the EGOK360 dataset as experimental results are sufficient to establish core concepts of rotation-invariant egocentric activity recognition.

\noindent\textbf{Experiment Configuration: } VIT360 training follows a similar approach we have seen in other related works~\cite{i3d, twostream, revisiting}. We choose Adam~\cite{adam} as our optimizer with the initial learning rate of $10^{-3}$, and StepLR as our learning rate scheduler for smooth training. We consider ten frames as one training input and six tangential patches as input discussed as $(T,n)$ in the method section. The experiments are conducted on three 2080Ti NVIDIA GPUs with a batch size of 16, both in the training and testing phase. The training and testing speed approaches 2 seconds/iterations and 1.5 seconds/iterations, respectively, achieving a throughput of nearly 80 and 106 frames per second. 

\noindent\textbf{Discussion on Activity Recognition Results: }Table ~\ref{tab:qtres} summarizes our experimental results on egocentric activity recognition on EGOK360 datasets. From these results, we can observe that the baseline method achieves marginally similar accuracy as reported in~\cite{egok360} on a fairly sampled new test set. Though these results are promising for controlled projections, the weaknesses of the baseline algorithm quickly appear when we perform random rotation during testing. Its accuracy drops significantly in both motion and spatial streams. However, considering the VIT360 for a similar effect, we obtain consistent results during fixed and random rotations. These achievements of VIT360 are useful considering the nature of 360$^\circ$ videos in the wild. In addition, we see a significant boost in performance using optical flow features computed using our inference techniques on 360$^\circ$ videos. The experimental results of VIT360 compared to the baseline on the same test set have shown an impressive performance boost, which provides strong evidence for the efficacy of VIT360 in egocentric activity recognition on 360$^\circ$ videos.

\noindent\textbf{Discussion on Optical Flow Results: } Based on results from Table ~\ref{tab:qtres}, the proposed optical flow inference technique is comparatively better than the traditional techniques. To address this, we also present qualitative results of our techniques and compare them with perspective flow-based techniques in Fig.~\ref{fig:floquality}. The qualitative results show more accurate optical flow for 360$^\circ$ videos using VIT360, which can be further explained by smooth displacement field, motion boundaries, and flow continuity around edges.
\section{Conclusion}
\label{sec:conclusion}
Omnidirectional flow estimation remains in its infancy because of the shortage of reliable benchmark datasets and tedious tasks dealing with inescapable radial distortions. This paper proposes the first perceptually natural-synthetic benchmark dataset, FLOW360, to close the gap, where comprehensive analysis shows excellent advantages over other datasets. Our dataset can be extended for other non-motion applications like segmentation and normal estimation task as well. Moreover, we introduce a siamese representation learning approach for omnidirectional flow (SLOF) instead of redesigning the convolution layer to adapt omnidirectional nature. Our method leverages the invariant rotation property of 360$^\circ$ videos to learn similar flow representation on various video augmentations. Meanwhile, we study the effect of different rotations on the final flow estimation, which provides a guideline for future work. Overall, the elimination of network redesigns aids researchers in exploiting existing architectures without significant modification leading to faster deployment in practical applications, such as egocentric activity recognition. 

Moreover, as a proof of concept, we propose VIT360 -- a vision transformer-based network pretrained with siamese representation - to achieve rotational invariance in 360$^\circ$ videos for egocentric activity recognition. VIT360 performs consistently regardless of the projection mode, achieving less than a 0.4\% accuracy gap whereas baseline methods drop in performance by 5\% to 14\% between fixed and random field-of-view projections. Our $360^\circ$ flow inference technique integrates seamlessly with VIT360 to boost the performance by almost 26\% (from 56.18 to 70.66), compared to the fixed projections trained networks. That is, VIT360 is more suitable for activity recognition in real-world scenarios thanks to its rotational invariant properties.


\bibliographystyle{IEEEtran}
\bibliography{egbib}

\clearpage

\setcounter{page}{1}

\appendices
\section{Flow Generator}

FLOW360 is created using Blender\footnote{\url{https://www.blender.org/}}, a free and open-source 3D creation suite. Blender provides an interface to write add-ons for automating workflows. We create Flow-generator, an add-on written for Blender-v2.92 to collect optical flow and several other data like depth information and normal maps. Flow-generator can be installed using the following steps:
\begin{enumerate}
    \item Download Flow-generator\footnote{\url{https://www.dropbox.com/s/v2mvjvs7ze8rzj1/flowgenerator.tar.gz?dl=0}} 
    \item In Blender-v2.92, follow:
    \begin{itemize}
        \item Edit $\rightarrow$ Preferences $\rightarrow$ Add-ons $\rightarrow$ Install
        \item Select flow\_generator.py from the Flow-generator project folder
        \item Install Add-on
        \item Search FlowGenerator
        \item Checkmark for installation
    \end{itemize}
    
    \item Setup compositor pipeline in Blender-v2.92
    \begin{itemize}
        \item Go to Compositing
        \item Select ``SetFlow Generator" from Custom Node Group
        \item If Custom Node Group is disabled press ``n" to make it visible
        \item Select the desired configuration and click ``SetEnv"
    \end{itemize}
    
    \item Camera Setup
    \begin{itemize}
        \item Go to Layout
        \item Open tab on the right side of the Layout view by pressing ``n" if not visible
        \item Go to ``CameraSetup" on the right tab
        \item Select the desired configuration and click ``Set Camera System"
    \end{itemize}
    
\end{enumerate}
Note that, we use cloud rendering to speed up the rendering process which requires crowd-render\footnote{\url{https://www.crowd-render.com/}} add-on. Flow-generator provides two basic functionality: (i) camera setup and (ii) compositor pipeline.

\subsection{Compositor Pipeline}
Compositor in Blender provides a pipeline to process render output. Flow-generator creates a basic pipeline (shown in~Fig. \ref{fig:flowgenerator}) to collect information like optical flow, images, depth maps, and so on. It also provides an easy way to cache the desired outputs in a structured format. Similarly, it also setups basic configurations like selecting render-engine and render passes. 
\subsection{Camera Setup}
\begin{figure*}[!t]
\centering
    \includegraphics[width=\linewidth]{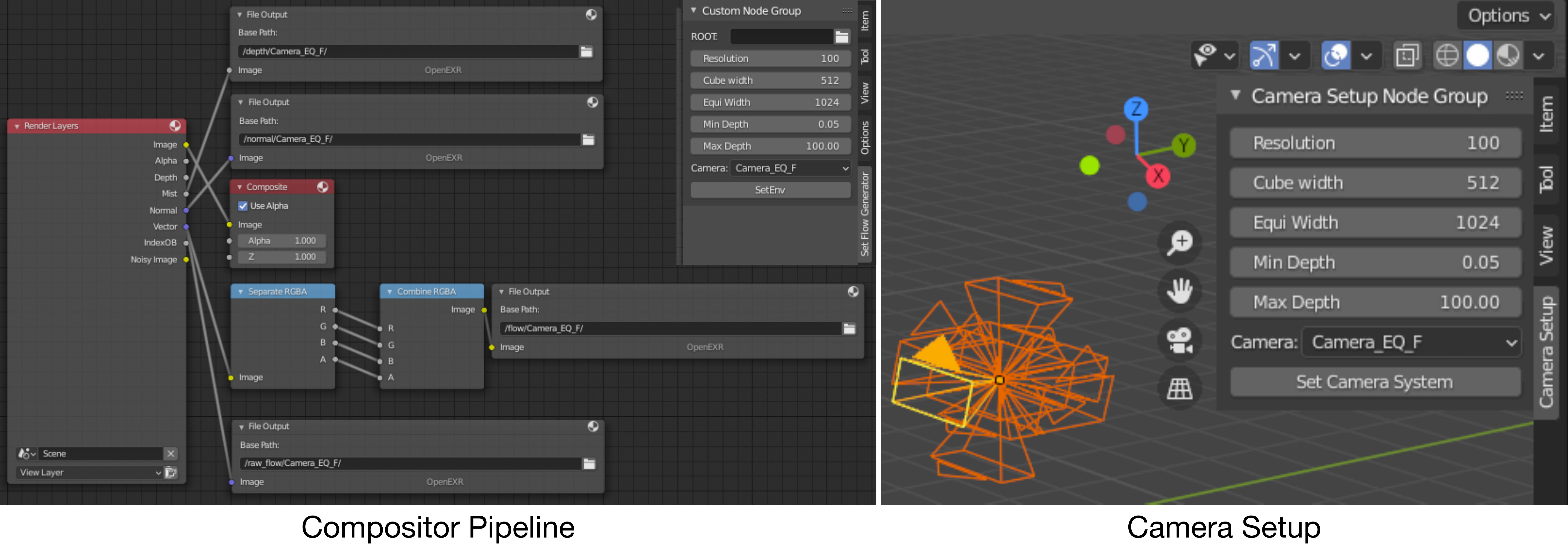}
    \caption{\textbf{FlowGenerator.} Flow-generator serves two major purposes: (\textbf{Left}) Setting up a compositor pipeline which involves setting up file paths, pipelining desired output, and setting initial configurations like render-engine and render passes. (\textbf{Right}) Setting up camera configurations and additional information like resolution, dimensions, and depth of the scene.}
    \label{fig:flowgenerator}
\end{figure*}

Camera Setup (shown in~Fig. \ref{fig:flowgenerator}) can be accessed via the 3D-Layout tab in Blender-v2.92 as suggested above. Camera Setup creates an omnidirectional camera with a full 360$^\circ$ field of view. The camera setup consists of twelve different cameras (six perspectives and six 360$^\circ$) out of which any omnidirectional camera (default Camera\_EQ\_F) can be selected as the main camera for rendering omnidirectional videos. It also provides an interface to adjust configurations like resolution, dimension, and depth of the scene.

\section{Experimental Setting and More Evaluation for Optical Flow Estimation}
\label{sec:experiment}
We conduct our experiment in PyTorch (1.9.0+cuda10.2) using the version of Python (3.8.10). Additional environment detail is provided in project\footnote{\url{https://www.dropbox.com/s/a6qioejg6yrxo7s/SLOF.tar.gz?dl=0}}.

Following is the list of configurations we used for our project:
\begin{itemize}
    \item Train/Val Batch Size: Ours(16/12 - 8Gpus), Finetune(8/12 - 4 Gpus)
    \item Number of iterations on RAFT: 12
    \item Loss: CosineSimilarity, Optical Flow L1-loss
    \item Optimizer: AdamW
    \item Scheduler: OneCycleLR
    \item EarlyStopping: Patience (5) and Min-delta (1e-4)
    \item Others: Gradient Clipping, GradScaler
    \item Dataset: FLOW360\footnote{\url{https://www.dropbox.com/s/nvzhazq99bg46f2/FLOW360_train_test.zip?dl=0}. Note that for better visualization please clip optical flow in the range of (-40, 40) or lower.}
\end{itemize}

We suggest our readers to refer sample videos\footnote{\url{https://www.dropbox.com/s/54mmjvoz6844mci/trailer.mp4?dl=0}}$^,$\footnote{\url{https://www.dropbox.com/s/gvihrzj528d92uj/videos.zip?dl=0} We recommend to use VLC-Media player to play these videos.} for demo purpose.

\subsection{Distortion Density Map}
\label{sec:distortion}
We compute distortion mask (U$_d$, D$_d$, F$_d$, B$_d$, R$_d$, L$_d$) in a cube-map with six faces: Up(U), Down(D), Front(F), Back(B), Right(R) and Left(L) respectively. This distortion density cube-map projection is then projected to equirectangular projection using spherical coordinate transformation as
\begin{equation}
  (x_s,y_s,z_s) = (\sin{\theta}\cos{\phi}, \sin{\theta}\sin{\phi},\cos{\theta}).
\end{equation}

To compute density mask (C$_d$, where C$\in$(U, D, F, B, R, L)) in each face, we define a meshgrid for co-oridnates $x$ and $y$ ranging from $([-1,1])$ with dimension of $(256,256)$. The coordinates $(x,y)$ are used to compute a radius map (r) of size $(256,256)$ as shown in Eq.~\eqref{eq:radius}.

\begin{equation}
  r = \sqrt{x^2 + y^2}
  \label{eq:radius}
\end{equation}

In our paper, we have shown that the radial distortion is higher towards the center in the polar region which corresponds to (U, D) faces of cube-map projections. Similarly, in the equatorial regions i.e., the rest of the faces (F, R, B, L) show a higher distortion rate away from the center. We compute two distinct distortion maps (C$_d$), one for polar regions (U, D) and another for equatorial regions (F, B, R, L) as shown in Eq.~\eqref{eq:compute}

\begin{equation}
\label{eq:compute}
C_d =
\begin{cases} 
      1 - r/\max(r) & \text{if, C $\in$ \{U,D\} }, \\
      r/\max(r) & \text{otherwise}
   \end{cases}
\end{equation}

Please refer to code\footnote{\url{https://www.dropbox.com/s/q1d4eoqvj2a30ij/distortion_weight.ipynb?dl=0}} for additional details.

\subsection{Impact of Rotational Invariance}
We perform additional experiments to quantify the impact of rotational invariance (see table below, Table.\ref{tab:rotinv}). The improvement is noticeable as seen in LiteFLowNet360~\cite{revisiting} and Tangent Images~\cite{tangent}-based optical flow estimation with rotational invariance.

\begin{table}[ht]
\small
\caption{Impact of rotational invariance}
\centering
\begin{tabular}{|c|c|c|}
\hline
Method & w/o Rot.Inv (EPE) & w Rot.Inv (EPE) \\
\hline
LiteFLowNet360~\cite{revisiting} & 3.95 & \textbf{2.52} \\
Tangent Images~\cite{tangent} & 3.57 & \textbf{1.78} \\
\hline
\end{tabular}
\label{tab:rotinv}
\end{table}

\begin{figure*}[!t]
\centering
    \includegraphics[width=\textwidth]{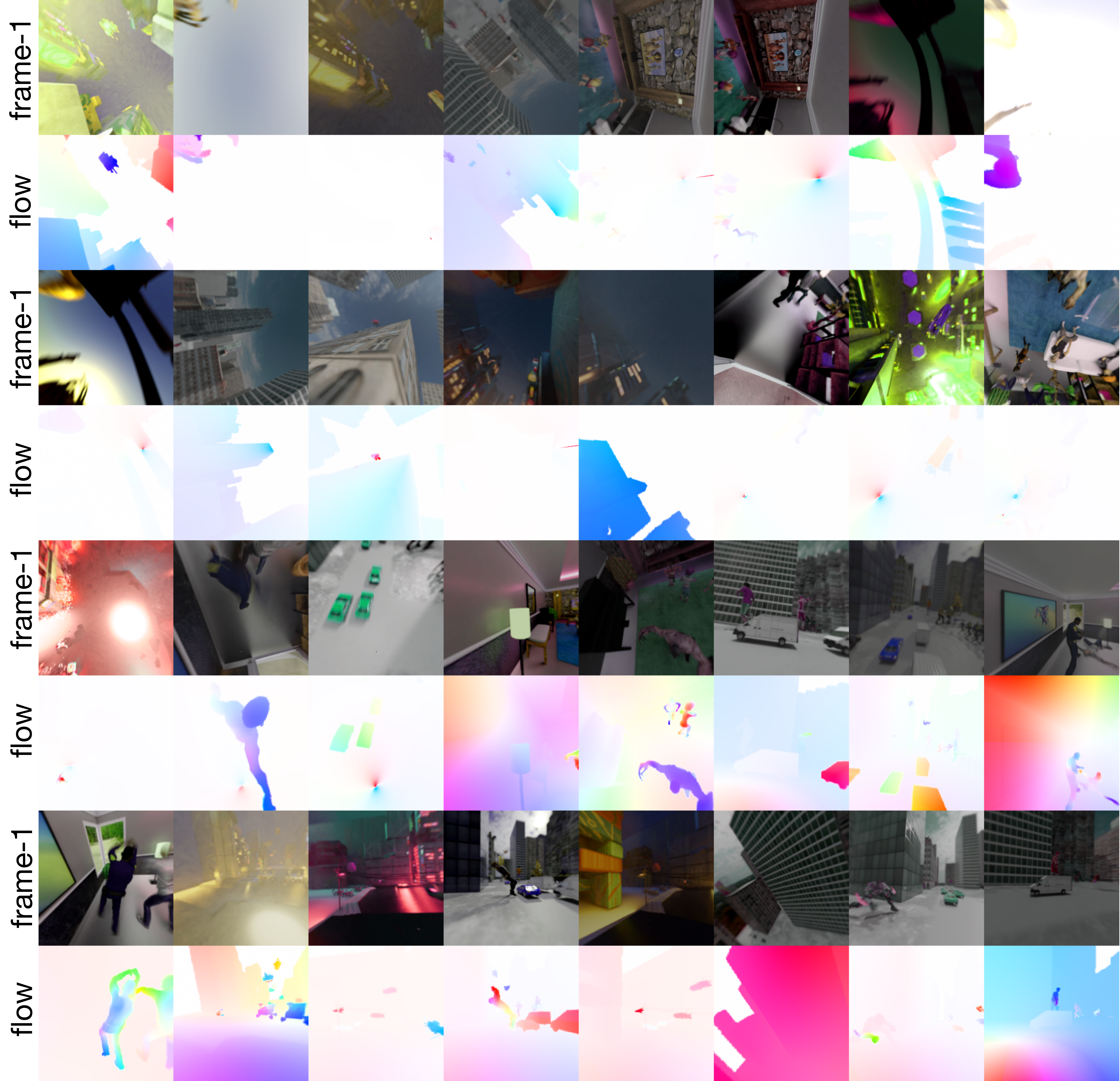}
    \caption{\textbf{More Motion and Scene Diversity - Train Set}. Illustrating motion and scene diversity via randomly sampled tangential plane from FLOW360 dataset. Flow360 contains a range of scene complexity governing varied properties like texture, illumination, human, building, cars, and other 3D assets. Similarly, the various level of motion complexity can be seen for similar scenes ranging from smaller to larger displacement. As explained in the main paper the dataset also contains other complexities like motion blur, camera focus/defocus, camera distortion, and environmental effects.}
    \label{fig:trainmont}
\end{figure*}

\begin{figure*}[!t]
\centering
    \includegraphics[width=\textwidth]{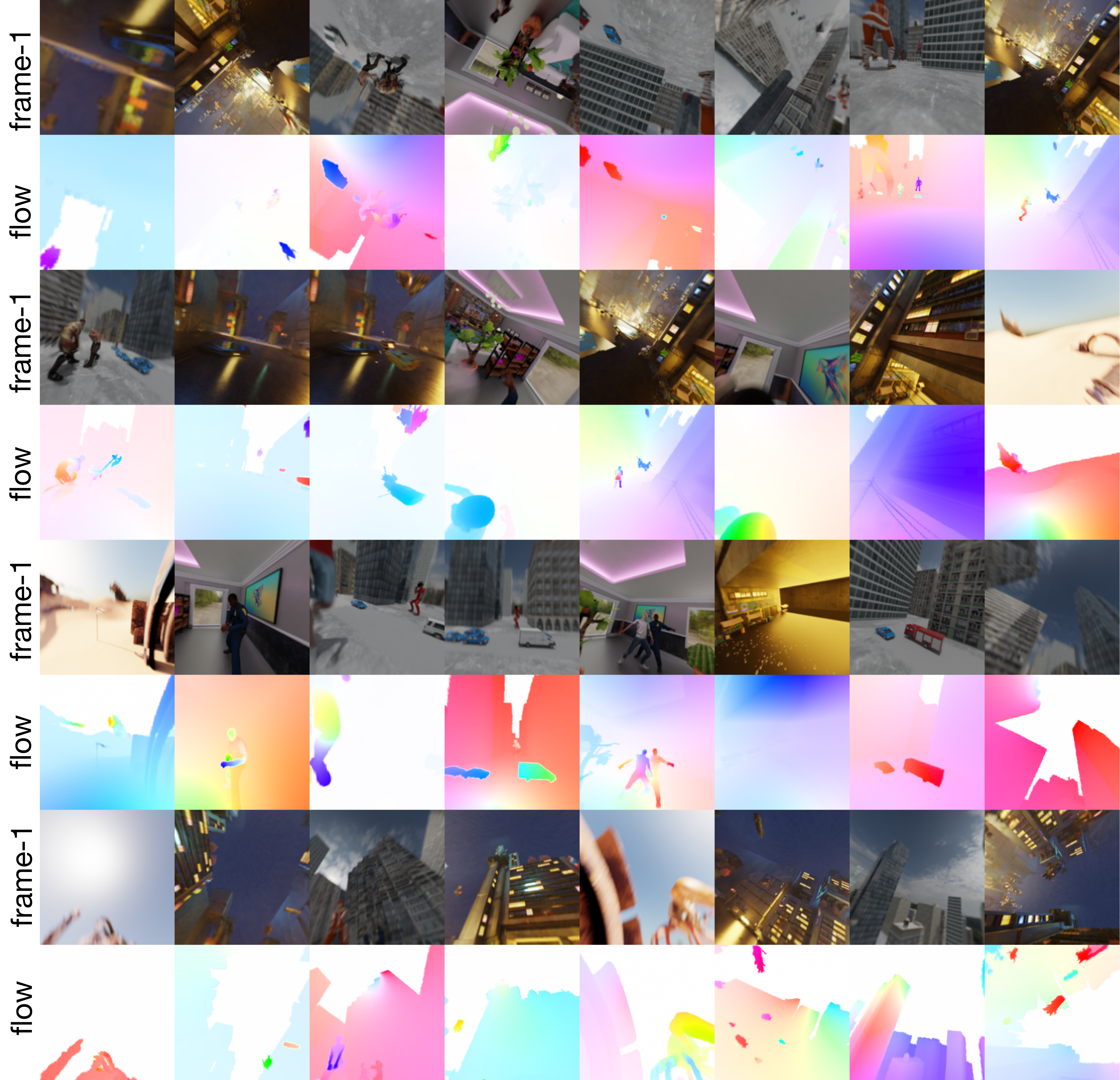}
    \caption{\textbf{More Motion and Scene Diversity - Test Set}. Illustrating motion and scene diversity via randomly sampled tangential plane from FLOW360 dataset. Flow360 contains a range of scene complexity governing varied properties like texture, illumination, human, building, cars, and other 3D assets. Similarly, the various level of motion complexity can be seen for similar scenes ranging from smaller to larger displacement. As explained in the main paper the dataset also contains other complexities like motion blur, camera focus/defocus, camera distortion, and environmental effects.}
    \label{fig:testmont}
\end{figure*}


\end{document}